\pdfoutput=1
\documentclass[10pt,twocolumn,letterpaper]{article}
\usepackage{cvpr}
\usepackage{times}
\usepackage{graphicx}
\usepackage{amsmath}
\usepackage{amssymb}
\usepackage{mathtools}
\usepackage{xcolor}
\usepackage{float}
\usepackage{booktabs}
\usepackage{color}
\usepackage[caption=false]{subfig}

\usepackage[pagebackref=true,breaklinks=true,letterpaper=true,colorlinks,bookmarks=false]{hyperref}
\newcommand{\tabincell}[2]{\begin{tabular}{@{}#1@{}}#2\end{tabular}}
\newcommand{\parsection}[1]{\vspace{1mm}\noindent\textbf{#1:}~}

\newcommand{\reals}{\mathbb{R}}

\DeclareMathOperator*{\argmin}{arg\,min}

\cvprfinalcopy 

\def\cvprPaperID{5197} 

\usepackage{capt-of,etoolbox}
\makeatletter
\apptocmd\@maketitle{{\introfig{}\par}}{}{}
\makeatother

\ifcvprfinal\fi
\begin{document}

\title{Generating Masks from Boxes by\\Mining Spatio-Temporal Consistencies in Videos}

\author{Bin Zhao \quad Goutam Bhat \quad  Martin Danelljan \quad  Luc Van Gool \quad  Radu Timofte\\
Computer Vision Lab, D-ITET, ETH Zurich, Switzerland\\
{\tt\small \{bzhao,goutam.bhat,martin.danelljan,vangool,radu.timofte\}@ethz.ch}
}

\newcommand{\introfig}{\centering\vspace{-6mm}
\includegraphics[width=0.95\textwidth]{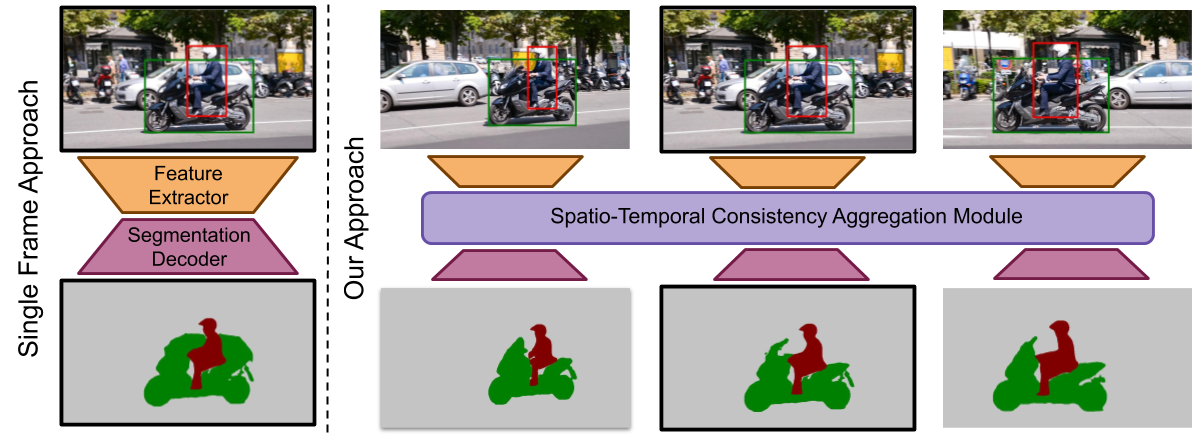}\vspace{-0.5mm}
  \captionof{figure}{When only using a single frame, predicting object masks from bounding boxes often leads to failures (left), since the outline of an object is difficult to resolve. By using video, our approach aggregates information across several frames. In this example, it identifies the car as background through the neighboring frames, while the scooter remains within the box, allowing it to be accurately segmented.}
\label{fig:intro}\vspace{5mm}}

\maketitle

\begin{abstract}
Segmenting objects in videos is a fundamental computer vision task. The current deep learning based paradigm offers a powerful, but data-hungry solution. However, current datasets are limited by the cost and human effort of annotating object masks in videos. This effectively limits the performance and generalization capabilities of existing video segmentation methods. To address this issue, we explore weaker form of bounding box annotations.

We introduce a method for generating segmentation masks from per-frame bounding box annotations in videos.
To this end, we propose a  spatio-temporal aggregation module that effectively mines consistencies in the object and background appearance across multiple frames. We use our resulting accurate masks for weakly supervised training of video object segmentation (VOS) networks. We generate segmentation masks for large scale tracking datasets, using only their bounding box annotations. The additional data provides substantially better generalization performance leading to state-of-the-art results in both the VOS and more challenging tracking domain.

\end{abstract}

\section{Introduction}
Segmenting objects in videos is an important but challenging task with many applications in autonomous driving \cite{ros2015vision, saleh2016kangaroo}, surveillance \cite{cohen1999detecting, erdelyi2014adaptive} and video editing. The field has been driven by the astonishing performance of deep learning based approaches \cite{bhat2020learning,oh2019video,voigtlaender2019feelvos}. However, these methods require large amount of training images with pixel-wise annotations. Manually annotating segmentation masks in videos is an extremely time-consuming and costly task.
Existing video datasets with segmentation labels \cite{Pont-Tuset_arXiv_2017,xu2018youtube_dataset} therefore do not provide the large-scale diversity desired in deep learning. This effectively limits the potential of current state-of-the-art approaches. 

To address this issue, it is tempting to consider weaker forms of human annotations. In particular, object bounding boxes are substantially faster to annotate. They provide horizontal and vertical constraints on the extent of the segmentation mask. Importantly, there already exist large scale video datasets with bounding box annotations \cite{fan2019lasot,Huang2019,muller2018trackingnet,ILSVRC15}. Hence, a method of effectively leveraging these annotations for video segmentation purposes would substantially enlarge the pool of available training data. Ideally, simply converting the video box annotations to object masks would allow existing video segmentation approaches to integrate these annotations using standard supervised techniques, not requiring any modification of losses or architectures. We therefore investigate the problem of generating object segmentations from box-annotations in videos.

Generating masks from box-annotated videos is a deceptively challenging task.
The background scene is often cluttered or contains similar objects. Objects can change appearance rapidly and often undergo heavy occlusions. Existing approaches \cite{luiten2018premvos, voigtlaender2020siam} only address the single frame case, where these ambiguities are difficult, or sometimes impossible to resolve due to the limited information.
However, the aforementioned problems can be greatly alleviated if we can utilize multiple frames in the video.
As the object moves relative to the background, we can find consistencies over several example views of the object and background. While object regions should consistently stay inside the box, background patches can move from inside to outside the object box over the duration of the video sequence. 
For instance, in Fig.~\ref{fig:intro} the single frame approach fails to properly segment the scooter due to the background car. In contrast, our video-based approach can identify the car as background in earlier and later frames while the scooter is consistently within the box for all frames. The car is therefore easily excluded from the final segmentation in all frames.

Effectively exploiting the information encoded in the temporal information is however a highly challenging problem. 
Since the object and background moves and changes in each frame, standard fusion operations cannot extract the desired consistencies and relations. 
Instead, we propose a spatio-temporal aggregation module by taking inspiration from the emerging direction of deep declarative networks \cite{deepdeclarative}. Our module is formulated as an optimization problem that aims to find the underlying object representation that best explains the observed object and background appearance in each frame. It allows our approach to mine spatio-temporal consistencies by jointly reasoning about all image patches in all input frames. The resulting mask embeddings for each frame are then processed by a decoder to generate the final segmentation output.

\parsection{Contributions}
Our main contributions are as follows.
\textbf{(i)} We propose a method for predicting object masks from bounding boxes in videos. To the best of our knowledge, we are the first to address this problem. \textbf{(ii)} We develop a spatio-temporal aggregation module that effectively mines the object and background information over multiple frames. \textbf{(iii)} Through an iterative formulation, we can further refine the masks through a second aggregation module. \textbf{(iv)} We utilize our method to annotate large-scale tracking datasets with object masks, which are then employed for  Video Object Segmentation (VOS).
\textbf{(v)} We perform extensive experiments, demonstrating the effectiveness of our approach in the limited data domain. Moreover, we show that the data generated by our approach allows VOS methods to cope with challenges posed by the tracking setting, achieving state-of-the-art performance on both domains. Code, models, and generated annotations will be made available at \url{https://github.com/visionml/pytracking}.


\section{Related work}
\parsection{Semi-supervised video object segmentation}
Semi-supervised video object segmentation (VOS) is the task of classifying all pixels in a video sequence into foreground and background, given a target ground-truth mask in the first frame. A number of different approaches have been proposed for VOS in recent years, including detection based methods~\cite{caelles2017one, maninis2018video,voigtlaender2017online}, propagation based approaches~\cite{khoreva2017lucid, li2018video, perazzi2017learning, wug2018fast}, feature matching techniques~\cite{chen2018blazingly, hu2018videomatch, oh2019video, voigtlaender2019feelvos}, and meta-learning based methods~\cite{behl2018meta, bhat2020learning, robinson2020learning}. A crucial factor that has enabled the recent advancements in VOS has been the release of high quality datasets such as DAVIS \cite{Pont-Tuset_arXiv_2017} and YouTube-VOS \cite{xu2018youtube_dataset}. However, due to the high costs of performing pixel-wise mask annotations, VOS datasets are still small in size compared to other computer vision fields such as detection, object tracking \etc. Consequently, VOS methods often resort to generating synthetic videos using multiple image segmentation datasets in order to obtain more training samples~\cite{oh2019video,wug2018fast}.

\parsection{Weakly supervised segmentation}
Due to the high cost of collecting pixel-wise labels, different types of weak annotations have been utilized to guide segmentation tasks recently, such as image-level supervision \cite{ahn2018learning, huang2018weakly, kolesnikov2016seed, pinheiro2015image, zhou2018weakly}, points \cite{bearman2016s,maninis2018deep}, scribbles \cite{lin2016scribblesup, vernaza2017learning} and bounding boxes \cite{dai2015boxsup, hsu2019weakly, khoreva2017simple, kulharia12356box2seg, papandreou2015weakly}. Our work is more related to bounding box supervised segmentation. In \cite{dai2015boxsup, khoreva2017simple}, pseudo segmentation masks for training are generated using GrabCut \cite{rother2004grabcut} and the MCG proposals \cite{pont2016multiscale}. 
Recent work~\cite{hsu2019weakly} designs a multiple instance learning (MIL) loss by leveraging the tightness property of bounding boxes. In \cite{kulharia12356box2seg}, a per-class attention map is predicted to guide the cross entropy loss and avoid propagating incorrect gradients from the background pixels inside the bounding box. As opposed to introducing specialized losses, our method directly generates pseudo masks from boxes, which can then be used to train models. Importantly, the masks are generated by utilizing spatio-temporal consistencies across video frames, leading to improved segmentation accuracy.

\begin{figure*}[t]
\centering
  \includegraphics*[width=1.0\linewidth,trim=0 5 0 5]{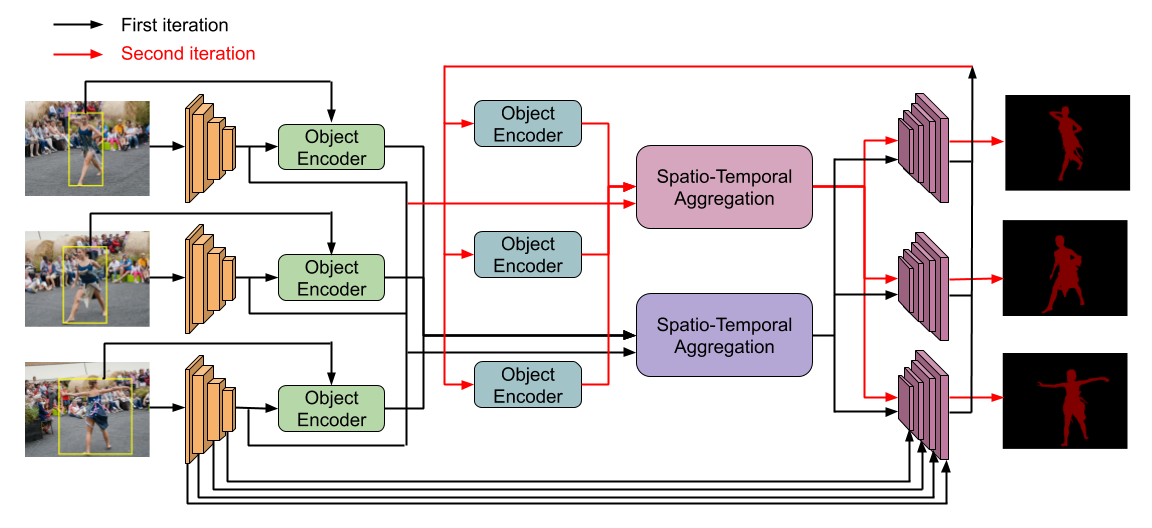}\vspace{-2mm}
  \caption{An overview of our architecture for segmenting an object from a box-annotated video. We first extract deep features from each frame. Features and boxes are then given to the object encoder (Sec.~\ref{sec:object-encoder}) to generate an object-aware representation. The spatio-temporal aggregation module (Sec.~\ref{sec:STA}) inputs object encodings and deep features from all frames. Its output is decoded to an object mask. We 
  refine the masks by iterating the process (Sec.~\ref{sec:iterative-model}) with a secondary object encoder and aggregation module to generate the final output.}\vspace{-3mm}
    \label{fig:box2seg architecture}
\end{figure*}

\parsection{Converting boxes to segmentation masks}
Generating a segmentation mask from the given object bounding box is an essential sub-task in instance segmentation, especially detection-based methods \cite{dai2016instance, hariharan2014simultaneous, he2017mask, li2017fully}. These approaches follow a multi-task learning strategy where a backbone network first extracts deep features and generates a set of proposals. Then detection and segmentation heads are used separately to predict an accurate bounding box and segmentation mask for the proposal. 
ShapeMask \cite{kuo2019shapemask} takes a bounding box detection as the initial shape estimate and refines it gradually, using a collection of shape priors. 
Luiten \etal~\cite{luiten2018premvos} train a modified DeepLabv3~\cite{Chen2018EncoderDecoderWA} model to output a mask, given a crop containing the object as input. In contrast to the previous approaches which only operate on single images, we address the task generating masks given box annotated videos as input. Our approach can exploit the additional temporal information in the video to predict more accurate masks, as compared to single image approaches. 

\parsection{Object Co-segmentation}
Object co-segmentation is the task of segmenting the common objects from a set of images. The concept of object co-segmentation was first introduced by Rother \etal in \cite{rother2006cosegmentation}, which minimizes an energy
function containing an MRF smoothness prior term and a histogram matching term. Subsequent work~\cite{rubinstein2013unsupervised} combines visual saliency and dense SIFT matching to capture the sparsity and visual variability of the common object in a group of images. 
Yuan \etal~\cite{yuan2017deep} introduce a deep dense conditional random field to automatically discover common information from co-occurrence maps generated by object proposals. The work \cite{li2018deep} integrates a mutual correlation layer into a CNN-based Siamese encoder-decoder architecture to perform co-segmentation. Similar to co-segmentation method, we segment objects using multiple images. However, our images are obtained from the same video. This enables exploiting the strong temporal consistency in videos to improve segmentation accuracy.

\section{Method}

We propose an end-to-end trainable architecture for the problem of segmenting an object in a video, given its bounding boxes in each frame. 
Our complete architecture is shown in Fig.~\ref{fig:box2seg architecture}. To fully exploit the temporal dimension, we aim to use detailed information of not only the target, but also the background context. Our backbone feature network $F$ therefore first separately encodes video frames $\{I_t\}_1^T$ containing the object as well as substantial background. The extracted deep features $x_t=F(I_t)$ along with the corresponding bounding box $b_t$ are given to the object encoder $B$, which provides an object-aware representation $e_t$ of each individual frame. By integrating the object bounding box, it provides information about hypothetical object and background regions. 

The object encodings and deep features $x_t$ from all frames are input to the spatio-temporal aggregation module $S$. The goal of this module is to generate an encoding of the object segmentation $s_t$ for each frame. The module $S$ aggregates appearance and box information from all frames and locations through an efficient and differentiable optimization processes. The iterative procedure fuses the different observations of the appearance $\{(x_t, e_t)\}$ by find an underlying representation of the object. This representation then generates the segmentation encoding $s_t$, which is processed by the segmentation decoder $D$ to predict preliminary object masks as $y_t = D(s_t, x_t)$. Our flexible architecture allows us to further improve the masks by feeding the results into a second spatio-temporal aggregation module, which predicts a set of refined segmentation encodings $\hat{s}_t$. The final segmentation masks $\hat{y}_t = D(\hat{s}_t, x_t)$ are then generated by the same decoder network. In the next section, we first detail our object encoder.

\subsection{Object Encoder}
\label{sec:object-encoder}
Accurately segmenting a specified object given only a single frame is a challenging problem. 
Since our goal is generic object segmentation, the type of object specified during inference may not even be represented in the training set. 
In general, it is therefore difficult to assess which image region inside a single bounding box belongs to the object in question. This is further complicated by cluttered scenes or presence of distractor regions in the background that are similar to the object itself. In these cases, determining object boundaries is particularly difficult. Moreover, multiple objects often overlap, making even the decision of which object to segment given the bounding box an ambiguous task.
All aforementioned problems are greatly alleviated and disambiguated if we can exploit several frames from a video sequence. As the object moves relative to the background, we can search for consistencies over several example views of the object appearance. While object regions should consistently stay inside the box, background patches can move from inside to outside the object box over the duration of the video sequence. This search for consistency is performed by our spatio-temporal aggregation $S$ through an iterative optimization process. It operates on information extracted from the individual frames by the object encoder, which we first detail in this section.

Directly extracting the segmentation from a single frame is difficult and prone to errors. We can however generate an object encoding from a single frame, capturing multiple \emph{possible} segmentation hypotheses. Each frame also gives detailed information about image patches, structures and patterns that are certainly not part of the object itself. These are regions of the image that are strictly \emph{outside} the bounding box, which provide important cues when combined with other frames in the sequence. To extract such frame-wise object information, we integrate an object encoder $B$. It takes information available in the frame $t$ by inputting the deep image representation $x_t = F(I_t)$ along with the object bounding box $b_t$. 
We first convert the bounding box $b_t$ to a corresponding rectangular mask representation in the input image coordinates. This is then processed by several convolutional and pooling layers, which increase the dimensionality while reducing the spatial resolution to be the same of the deep features $x_t$. The resulting activations are then concatenated with the features $x_t$ and further processed by several residual blocks.

Through the deep features $x_t$, the object encoder $B$ can extract candidate object shapes, which are used when searching for consistency over several frames. Specifically, the object encoder has three outputs,
\begin{equation}
    \label{eq:objenc}
    (e_t, w_t, m_t) = B(x_t, b_t) \,,\quad e_t, w_t, m_t \in \reals^{H \times W \times C} \,.
\end{equation}
All outputs have the same spatial resolution $H \times W$ as the deep features $x_t$. The abstract embedding $e_t$ holds information about the candidate object shapes and background regions in image $I_t$. Intuitively, at spatial location $(i,j)$, the activation vector $e_t[i,j] \in \reals^C$ encodes probable segmentations of the corresponding image region. Since the uncertainty of this information can vary spatially and over the feature channels, we also predict a corresponding confidence weight $w_t$. Indeed, regions outside the bounding box are certainly not part of the object, while regions inside the box can be ambiguous or difficult to interpret. The output $m_t$ corresponds to relevant single-frame object information that is directly given to the segmentation decoder $D$. The encoding $e_t$ and its confidence $w_t$ are given to the spatio-temporal aggregation module, detailed next.

\subsection{Spatio-Temporal Aggregation}
\label{sec:STA}
It is the task of the spatio-temporal aggregation module to mine the object information over multiple frames. However, designing a neural network module capable of effectively integrating information from multiple frames is a challenging and intricate problem as the object changes location and pose in each frame. As a result, temporal pooling, concatenation, or convolutions cannot find the desired consistencies. Moreover, these operations do not consider detailed global information. When deciding whether a patch corresponds to foreground or background, we need to find and reason about all similar patches in the given frames.

Our approach takes inspiration from the emerging direction of deep declarative networks~\cite{deepdeclarative}, which have shown promise in few-shot and meta-learning applications \cite{bhat2019learning, metaoptnet}, including video object segmentation \cite{bhat2020learning}. These approaches formulate the task of a neural network as an optimization problem. This problem is solved during the forward pass and allows end-to-end learning by ensuring that the solution to be differentiable \wrt all inputs. Note that our setting does not fall into the domain of few-shot or meta-learning. Yet, we demonstrate that the temporal aggregation task encountered in this work can be gracefully formulated as an optimization objective. We believe that our objective and its novel interpretation further widen the scope of applications for ideas stemming from deep declarative networks.  

The main idea of our formulation is to find an underlying object representation $z$ that best \emph{explains} the observed object embedding $e_t$. That is, the representation $z$ should indicate \emph{consistent} local correlations between the deep image features $x_t$ and the corresponding object embedding $e_t$. We formulate this as the problem of finding the best fitting local linear mapping from features vectors $x_t[i,j] \in \reals^{D}$ to corresponding embedding vectors $e_t[i,j] \in \reals^{C}$. This is most conveniently expressed as a convolution with the filter $z \in \reals^{K \times K \times D \times C}$, where $K$ is the kernel size. Using the squared error to measure the fit, our temporal aggregation module is formulated as
\begin{subequations}
    \label{eq:temporal-agg}
\begin{align}
    \label{eq:agg-module}
    &\{s_t\}_1^T = S\big(\{(x_t, e_t, w_t)\}_1^T\big) = \{x_t * z^*\}_1^T \quad \text{where} \\
    \label{eq:agg-obj}
    & z^* = \argmin_z  \frac{1}{T} \sum_{t=1}^T \big\|w_t \cdot (x_t * z - e_t)\big\|^2 + \lambda \|z\|^2.
\end{align}
\end{subequations}
The filter $z$ is thus optimized to predict the embedding $e_t$ from the features $x_t$. In order to minimize the objective, the filter must focus on consistent local correlations between $x_t$ and $e_t$, while ignoring accidental relations that do not reoccur. The predicted confidence $w_t$ actively weights the error at each spatio-temporal location and channel dimension through an element-wise multiplication. Our network can therefore learn to ignore information in $e_t$ that is deemed uncertain by predicting a low weight $w_t$, while emphasizing other information by giving a large importance weight. The regularization weight $\lambda$ is treated as a network parameter and thus learned during training.

During both inference and training, the optimization problem \eqref{eq:agg-obj} needs to be solved for every forward pass of the network. The solver thus needs to be efficient in order to ensure practical training and inference times. Moreover, the solution $z^*$ needs to be differentiable \wrt to the inputs $\{(x_t, e_t, w_t)\}_1^T$ and $\lambda$.
There exist approaches for directly back-propagation through the solution $z^*$ for convex problems such as \eqref{eq:agg-obj} using the implicit function theorem  \cite{deepdeclarative,metaoptnet} or closed-from expressions~\cite{r2d2}. However, these methods rely on accurately finding the optima $z^*$, which is not necessary in our case. Instead, we found the unrolled steepest-descent based optimization strategy \cite{bhat2019learning,bhat2020learning} to yield a simple and fast solution. As the algorithm employs iterative updates to $z$ through a differentiable closed-form expression, backpropagation is automatically achieved through the standard auto-differentiation implemented in popular deep learning libraries, such as PyTorch and TensorFlow.

After mining for spatio-temporal consistencies through the iterative minimization of \eqref{eq:agg-obj}, the filter $z^*$ contains a strong representation of the object. It encapsulates the consistent patterns and correlations of the object, integrating both spatial and temporal information. The output segmentation encoding $s_t$ of the spatio-temporal aggregation module is achieved by applying the optimized representation $z^*$ to the deep features $x_t$ of each frame in \eqref{eq:agg-module} as $s_t = x_t * z^*$. This is then input to our decoder $y_t = D(s_t, m_t, x_t)$, which generates a final object segmentation $y_t$.

\subsection{Iterative Refinement}
\label{sec:iterative-model}
In this section, we describe a method to further refine the object segmentations using existing components in our architecture.
The decoder module learns powerful segmentation priors by integrating deep features from different levels. It is able to extract accurate object boundaries and filter out potential errors. The segmentation embedding $s_t$ predicted by the spatio-temporal aggregation module \eqref{eq:temporal-agg} is thus enriched with these priors in order to generate the output segmentation $y_t$. Note that this represents new knowledge not seen by the aggregation module in the first pass. We can therefore utilize this information by feeding the output segmentation masks back into the aggregation step.

To this end, we create a secondary object encoder $\hat{B}$, taking the predicted segmentation $y_t$. Since the preliminary segmentation $y_t$ already encapsulates a detailed representation of the object extent, we found it to be sufficient for generating the object embedding $\hat{e}_t$ and confidence weights $\hat{w}_t$ used by the aggregation module. Thus for each frame we predict,
\begin{equation}
    \label{eq:objenc2}
    (\hat{e}_t, \hat{w}_t) = \hat{B}(y_t) \,,\qquad e_t, w_t \in \reals^{H \times W \times C} \,.
\end{equation}
Note that we do not re-generate the single-frame information $m_t$ later used by the decoder. Instead, we employ the one stemming from the original object encoder \eqref{eq:objenc}.

The object encoding $\hat{e}_t$ and corresponding weights $\hat{w}_t$ now include new and more accurate information about the object. We integrate this for mask prediction by inputting it to our spatio-temporal aggregation module \eqref{eq:temporal-agg} to generate new segmentation encodings $\{\hat{s}_t\}_1^T = S\big(\{(x_t, \hat{e}_t, \hat{w}_t)\}_1^T\big)$. Note that this implies solving a new optimization problem \eqref{eq:agg-obj}, which mines spatio-temporal consistencies. The final segmentation mask $\hat{y}_t$ is obtained using the same decoder module as $\hat{y}_t = D(\hat{s}_t, m_t, x_t)$. While the process could be repeated several times, we did not observe noticeable improvement from a third iteration. This is however expected as the strong segmentation priors of the decoder is already exploited by the aggregation module in the second iteration.

\subsection{Training}
\label{sec:training}
Our complete model is fully differentiable, and can therefore be trained end-to-end using existing video datasets annotated with segmentation masks. From a ground truth mask $y_t^\text{GT}$, we extract a corresponding bounding box $b_t$ by taking smallest axis-aligned box containing the mask $y_t^\text{GT}$. Our network is trained on sub-sequences of length $T$ by minimizing the loss,
\begin{equation}
    \label{eq:train-loss}
    L = \frac{1}{T} \sum_{t=1}^T \ell(y_t, y_t^\text{GT}) + \frac{1}{T} \sum_{t=1}^T \ell(\hat{y}_t, y_t^\text{GT}) \,.
\end{equation}
Here, $y_t$ and $\hat{y}_t$ is the segmentation output generated by the initial prediction and the refinement respectively. Further, $\ell$ denotes a generic segmentation loss.

For our experiments, we use the YouTube-VOS \cite{xu2018youtube_dataset} and DAVIS 2017 \cite{Pont-Tuset_arXiv_2017} datasets. In each epoch, we sample sequences from both datasets without replacement. Due to its larger size, we use a 6 times higher probability for YouTube-VOS 2019 compared to the DAVIS 2017 training set. We randomly sample sequences of length $T=3$ frames within a temporal window of length 100. For each frame, we first crop a patch that is 5 times larger than the ground-truth bounding box, while ensuring the maximal size to be equal to the image itself. We then resize the cropped patch to $832\times 480$ with the same aspect ratio. Only random horizontal flipping is employed for data augmentation. 

We initialize our backbone ResNet-50 network with Mask R-CNN weights from \cite{massa2018mrcnn}. All the remaining modules are initialized using \cite{he2015delving}. The network parameters are learned by minimizing Lovasz \cite{berman2018lovasz} segmentation loss. We use ADAM \cite{kingma2014adam} optimizer to update network parameters with a mini-batch size of 4. We train our network for 80k iterations with the backbone weights fixed. The learning rate is initialized as $10^{-2}$ and then reduced by a factor of 5 after 30k and 60k iterations. The entire training takes 32 hours on a single Nvidia TITAN Xp GPU.

\subsection{Implementation Details}
\label{sec:implementation}

\parsection{Architecture}
Here, we give further details about our architecture. 
We use a ResNet-50 backbone network as feature extractor. For the object encoder $B$ and spatio-temporal aggregation module $S$, we employ the third residual block and add another convolutional layer which reduces the dimensionality to $512$. The object encoder generates outputs \eqref{eq:objenc} with a dimension $C=16$.
We adopt the segmentation decoder used in \cite{bhat2020learning, robinson2020learning}. We first concatenate the segmentation embedding $s_t$ from \eqref{eq:temporal-agg} with the single-frame information $m_t$ from \eqref{eq:objenc}. The decoder then progressively increases the resolution while integrating deep features from different levels in $F$.
For the spatio-temporal aggregation module \eqref{eq:temporal-agg}, we first initialize the object representation $z$ to zero. We then apply 5 steepest descent iterations \cite{bhat2020learning} to optimize \eqref{eq:agg-obj} during training. The kernel size of $z$ is set to $K=3$.

\parsection{Inference}
In this section, we provide details about our inference procedure. For a given input video, we extract a sequence of $T$ frames. Since our method benefits from using \emph{different} views of the target and background, we do not extract directly subsequent frames as they are highly correlated. Instead, we take $T$ frames with an inter-frame interval of $\Delta$. In order to segment all frames, we simply proceed by shifting the sub-sequence one step each time. We generally employ $\Delta = 15$. We analyze the impact of the sequence length $T$ in Sec.~\ref{sec:ablation}. For the spatio-temporal aggregation \eqref{eq:train-loss}, we found it beneficial to increase the number of steepest-descent iterations to 15 during inference.

\begin{figure*}
    \centering
    \subfloat{
    \begin{minipage}[c]{0.12\textwidth}
    \includegraphics[width=\textwidth]{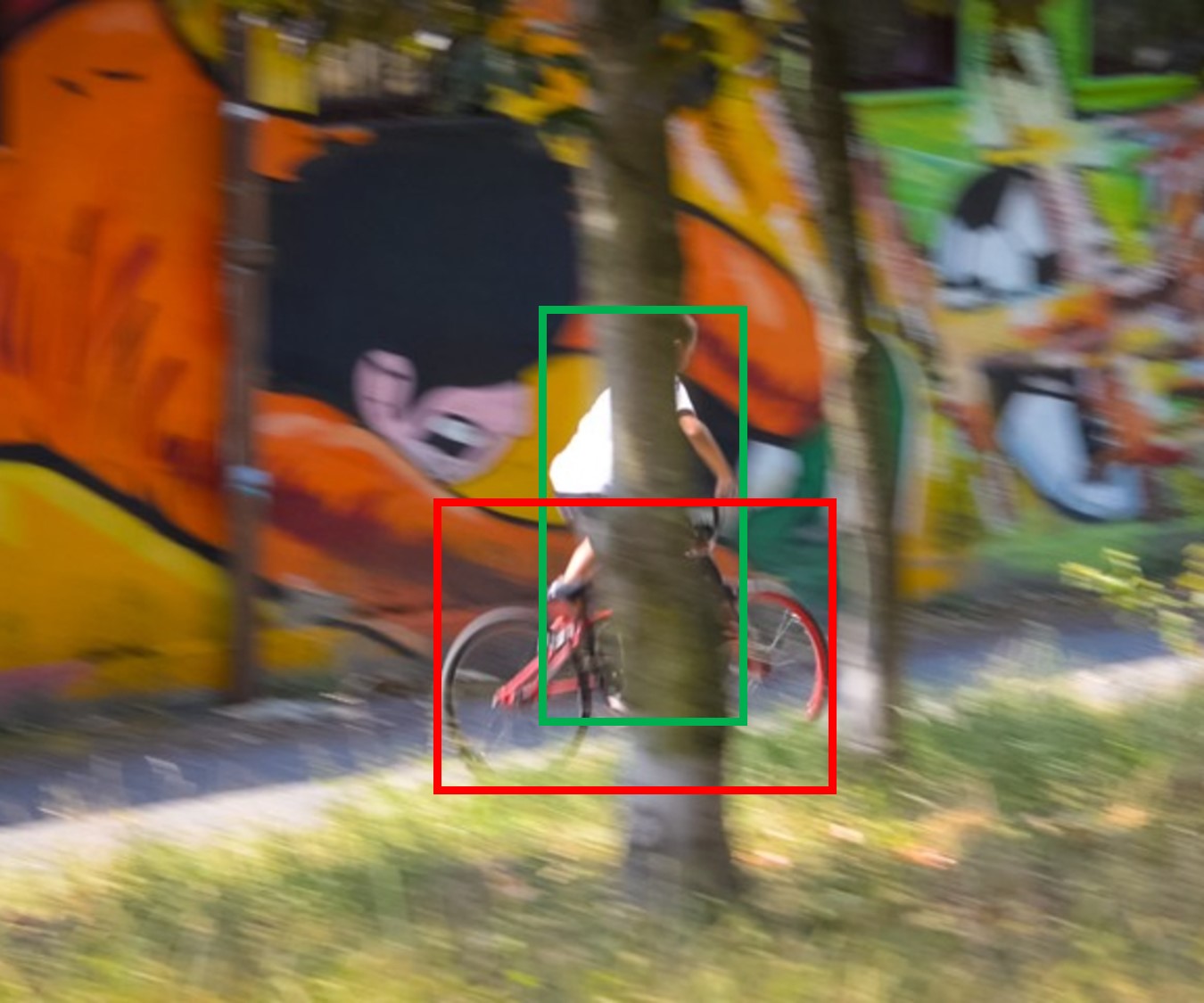}
    \includegraphics[width=\textwidth]{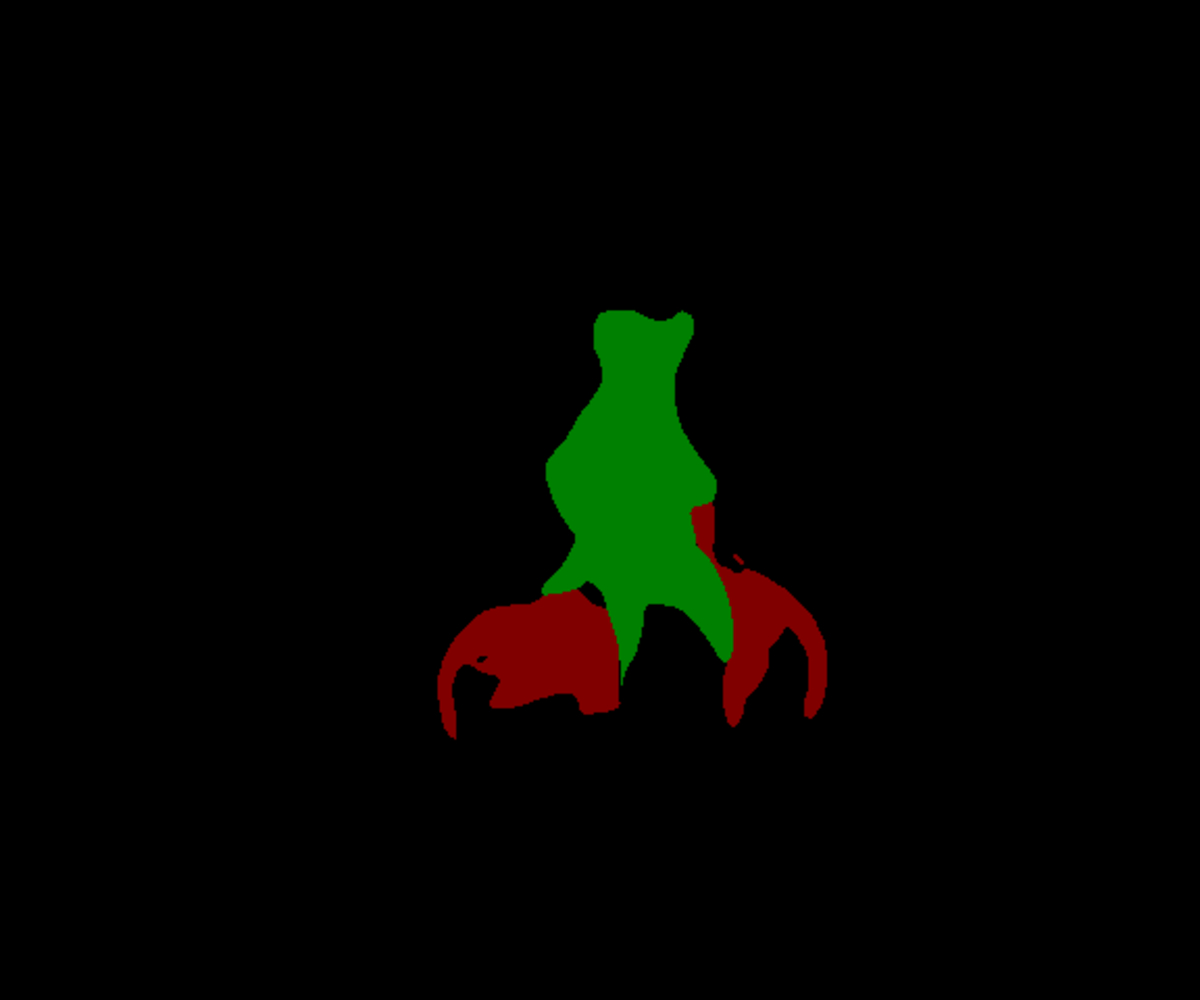}
    \includegraphics[width=\textwidth]{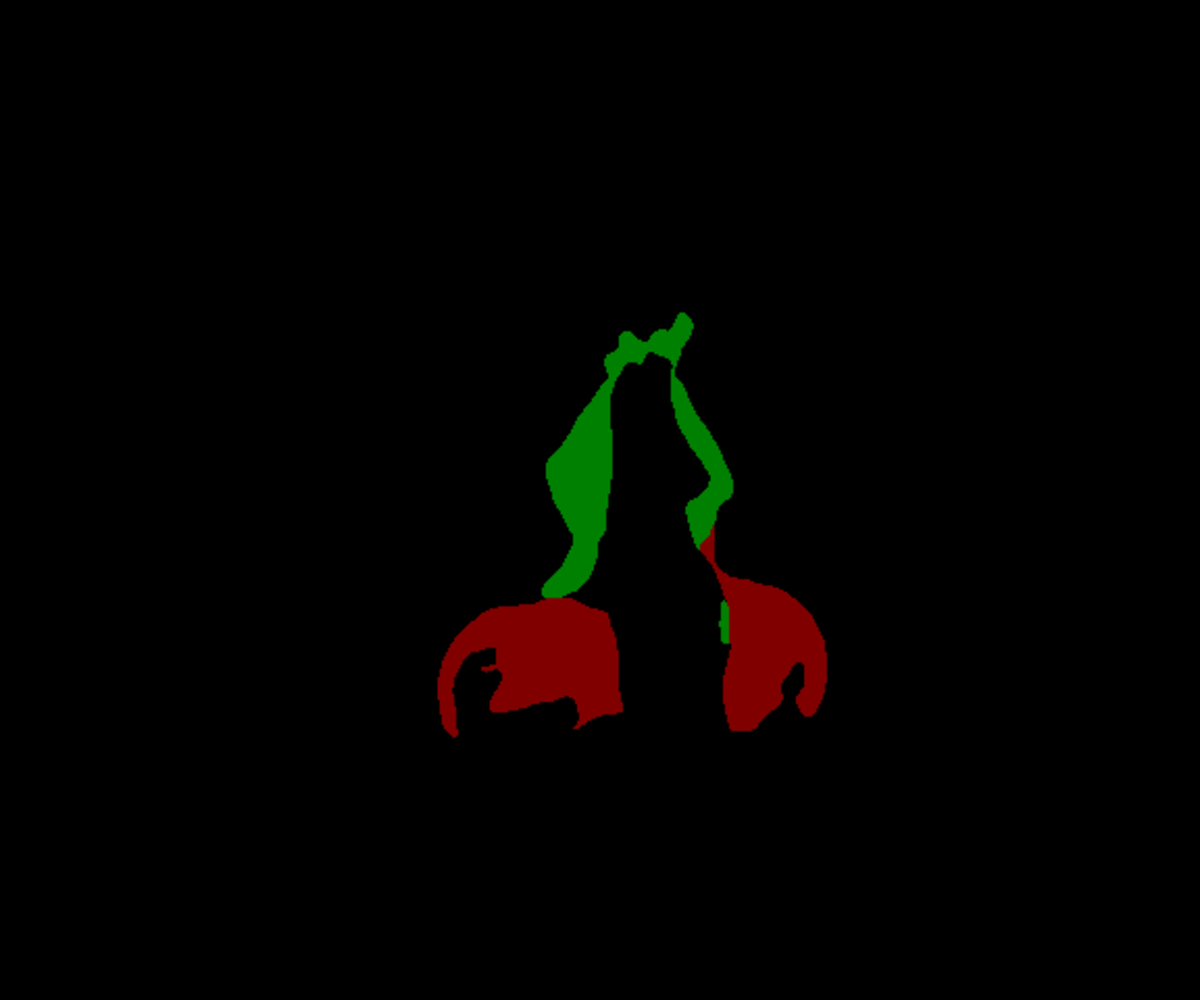}
	\end{minipage}
	}\hspace{-2mm}
	\subfloat{
    \begin{minipage}[c]{0.12\textwidth}
    \includegraphics[width=\textwidth]{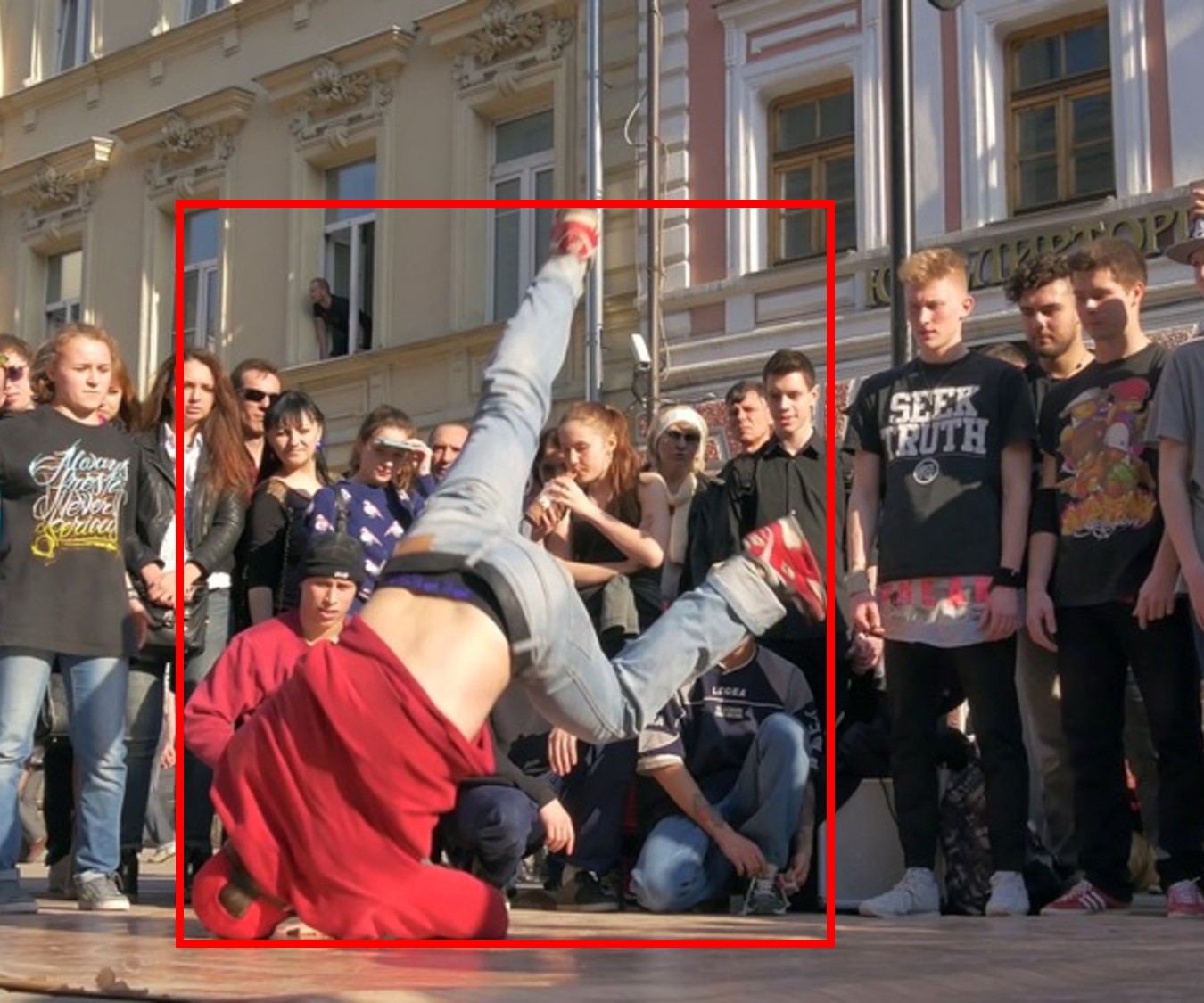}
    \includegraphics[width=\textwidth]{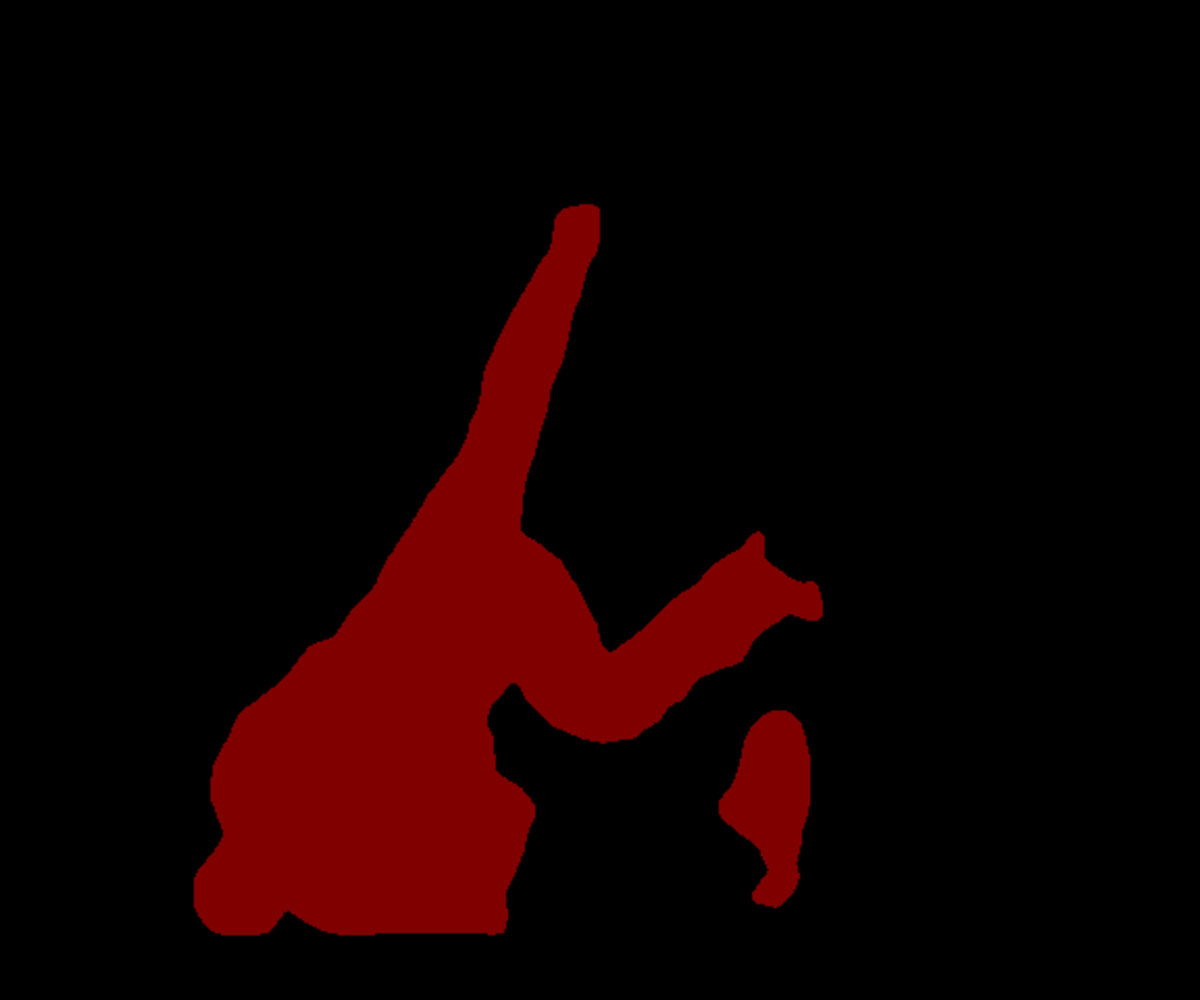}
    \includegraphics[width=\textwidth]{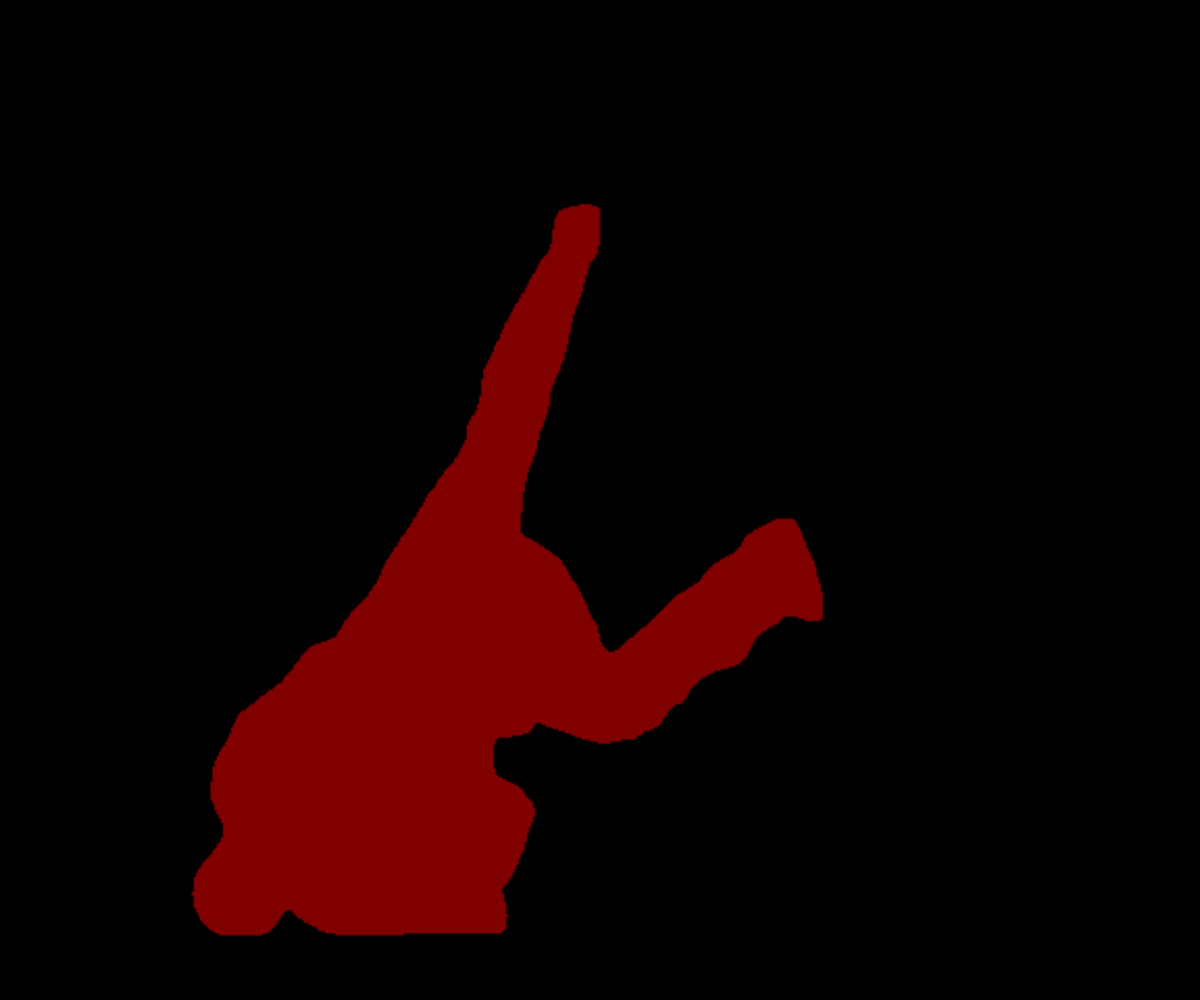}
	\end{minipage}
	}\hspace{-2mm}
	\subfloat{
    \begin{minipage}[c]{0.12\textwidth}
    \includegraphics[width=\textwidth]{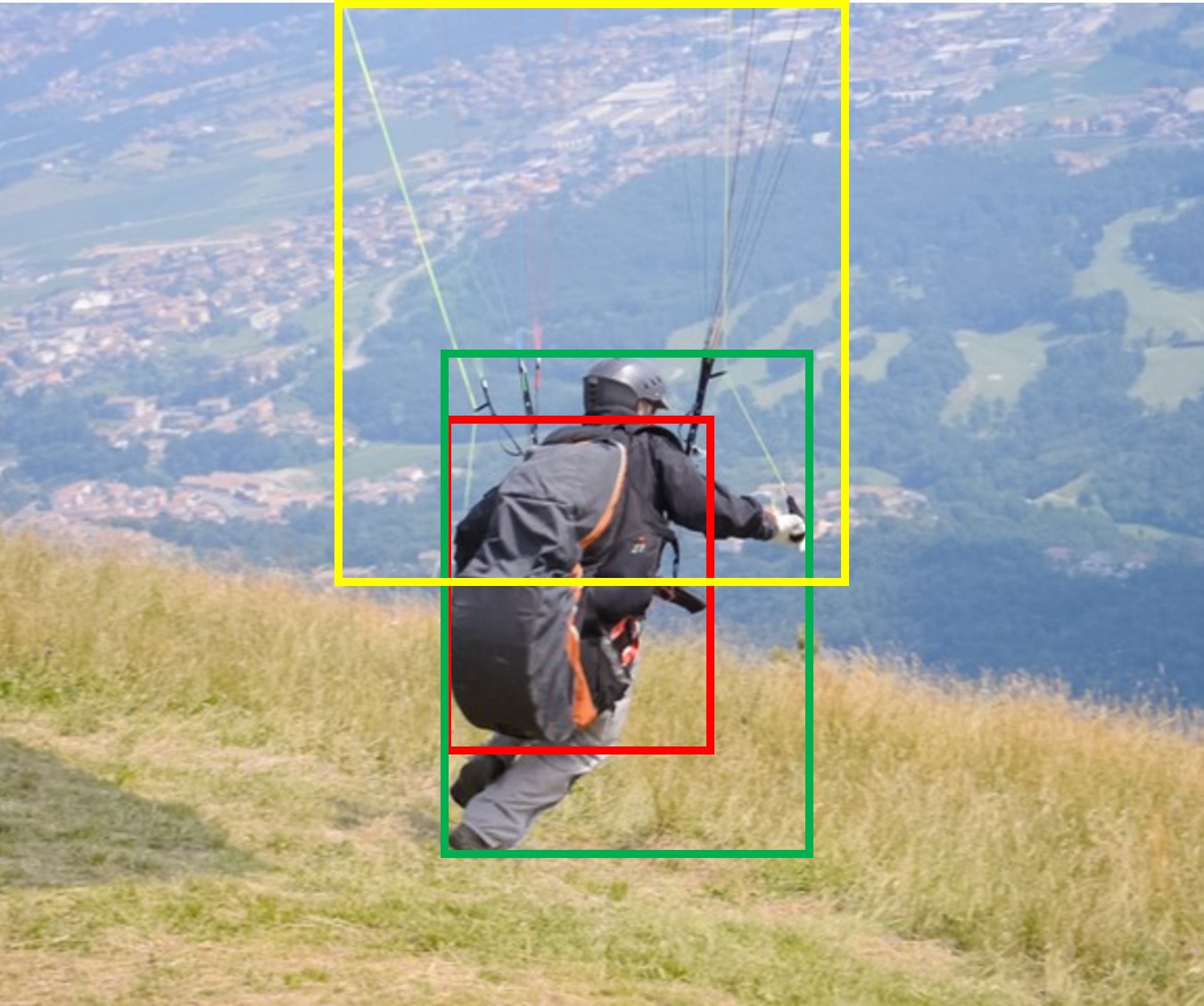}
    \includegraphics[width=\textwidth]{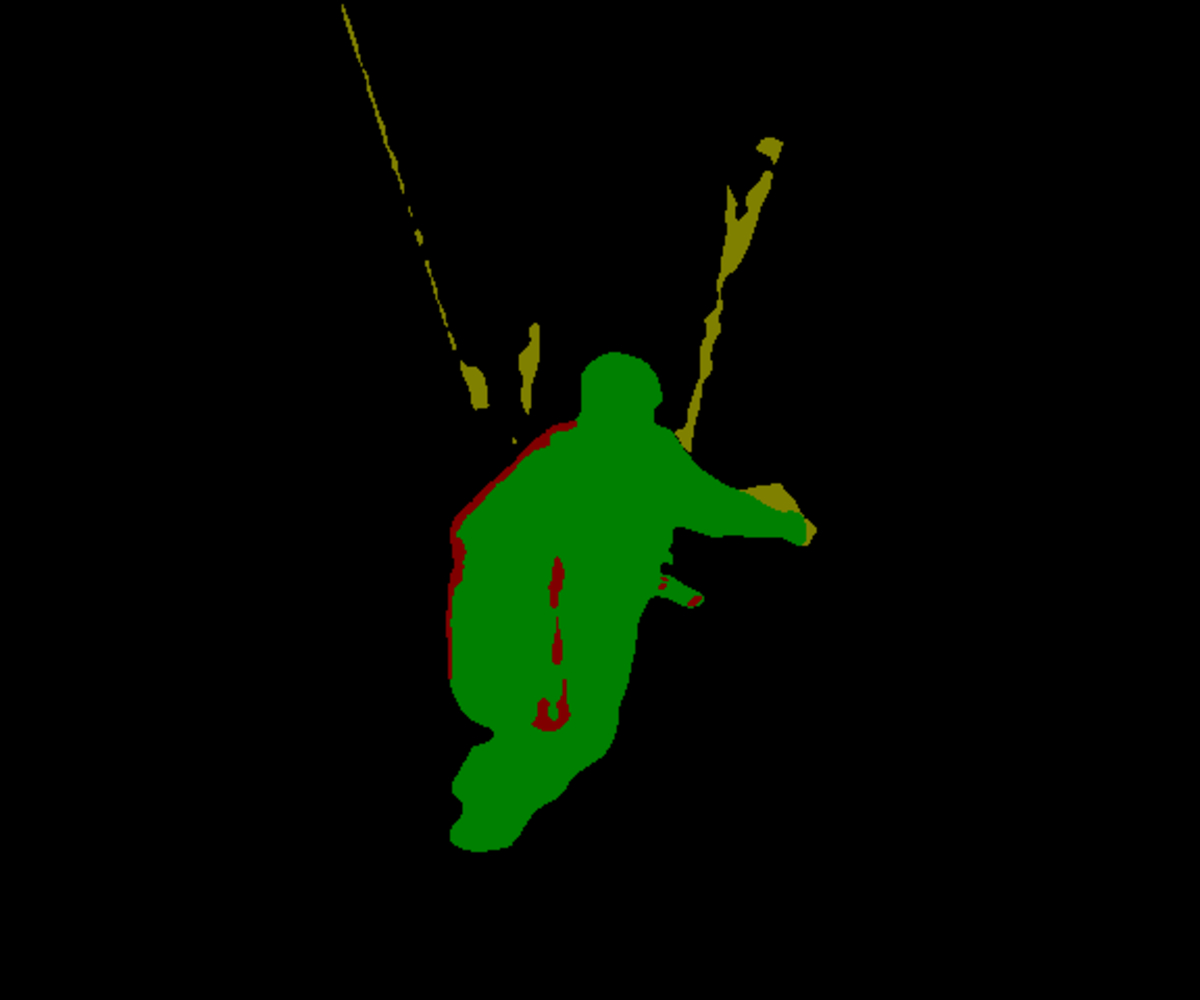}
    \includegraphics[width=\textwidth]{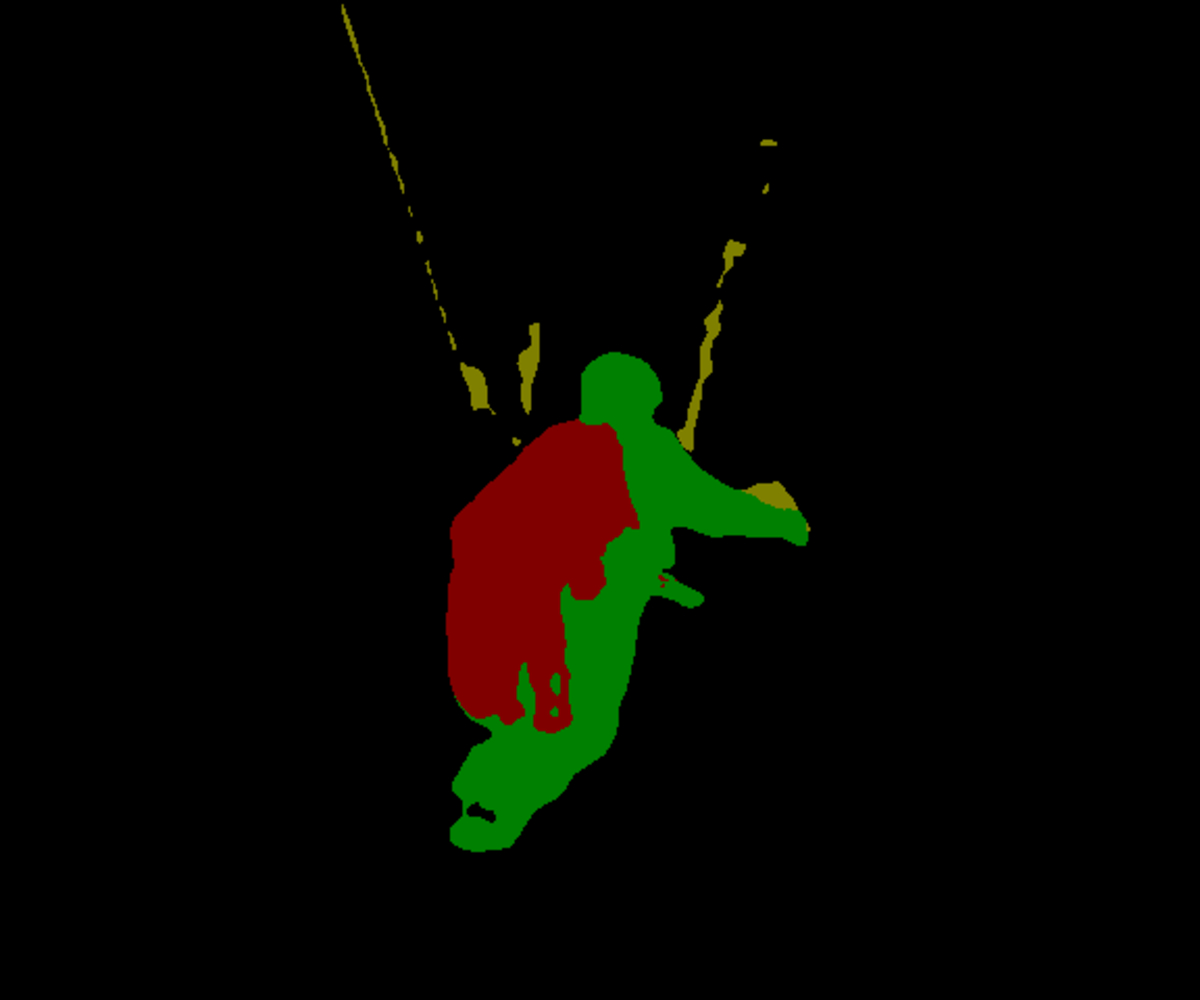}
	\end{minipage}
	}\hspace{-2mm}
	\subfloat{
    \begin{minipage}[c]{0.12\textwidth}
    \includegraphics[width=\textwidth]{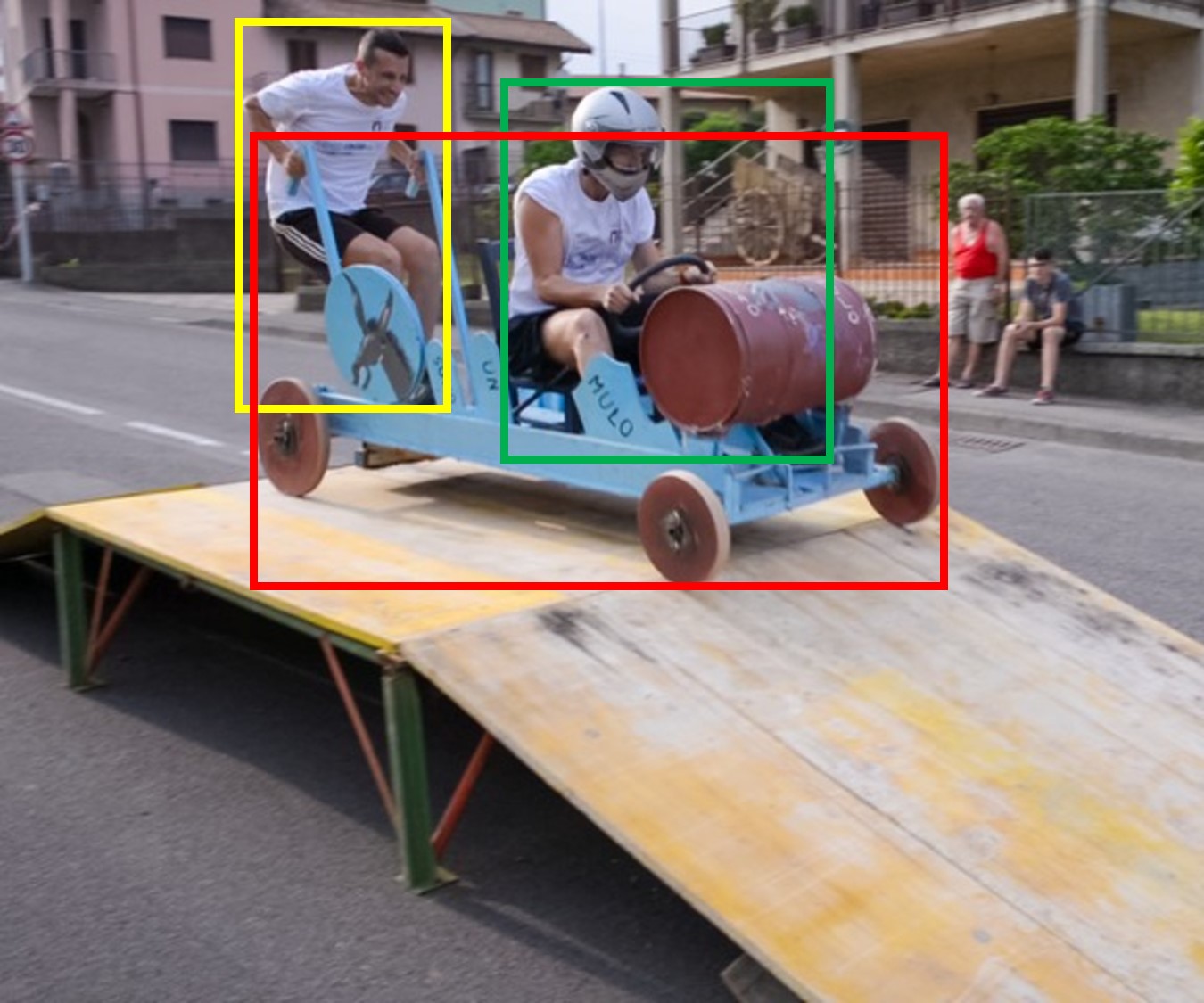}
    \includegraphics[width=\textwidth]{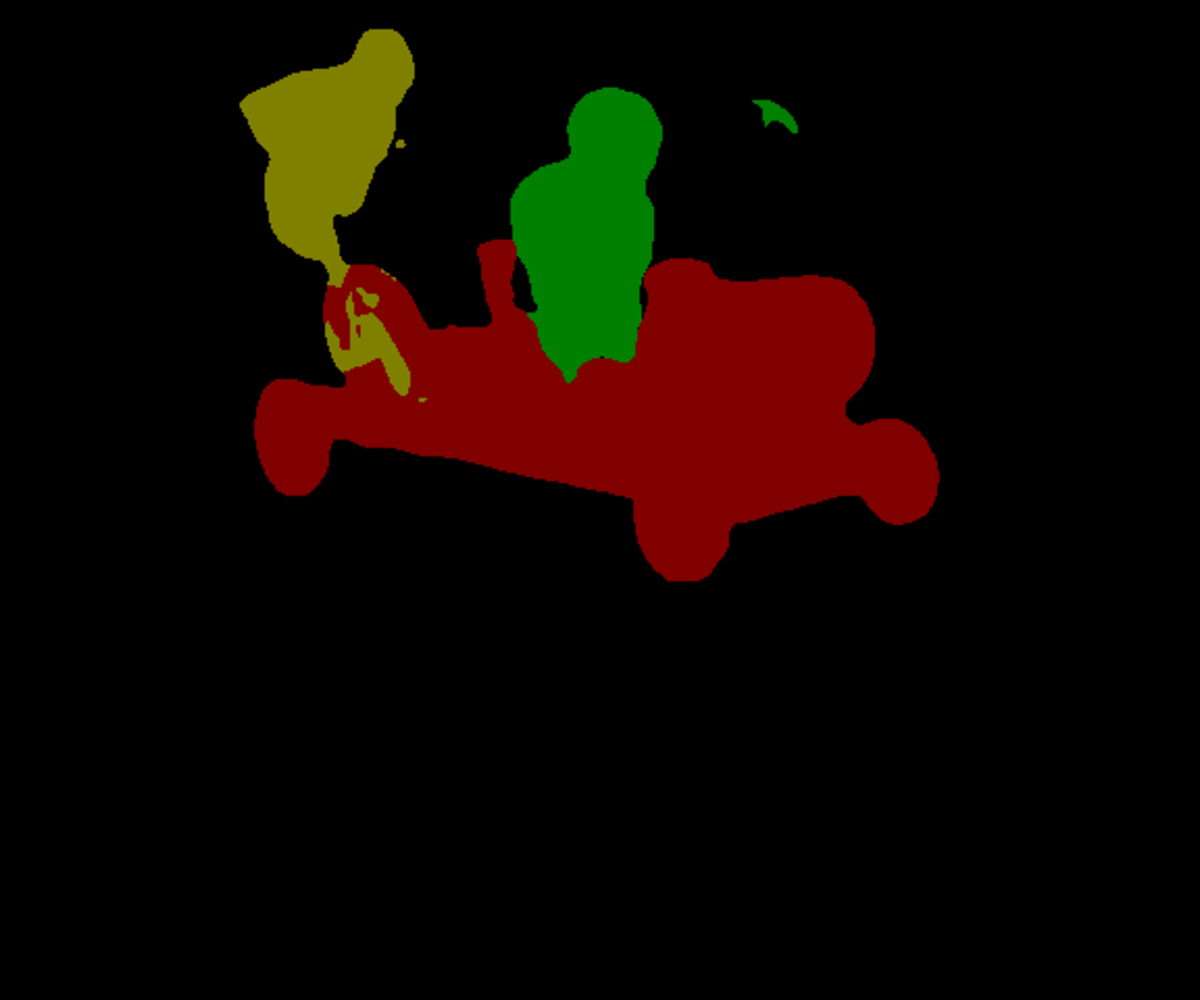}
    \includegraphics[width=\textwidth]{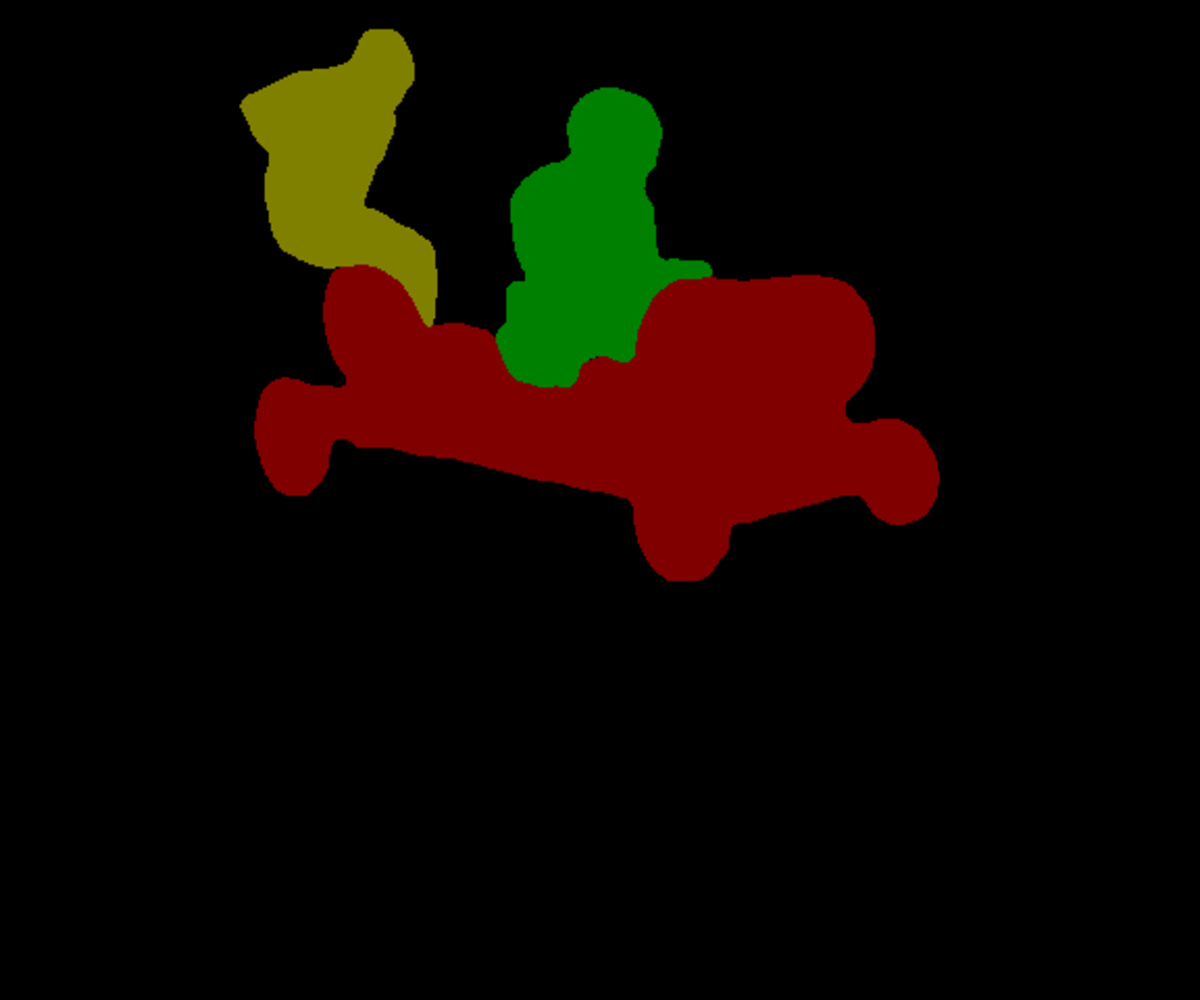}
	\end{minipage}
	}\hspace{-2mm}
	\subfloat{
    \begin{minipage}[c]{0.12\textwidth}
    \includegraphics[width=\textwidth]{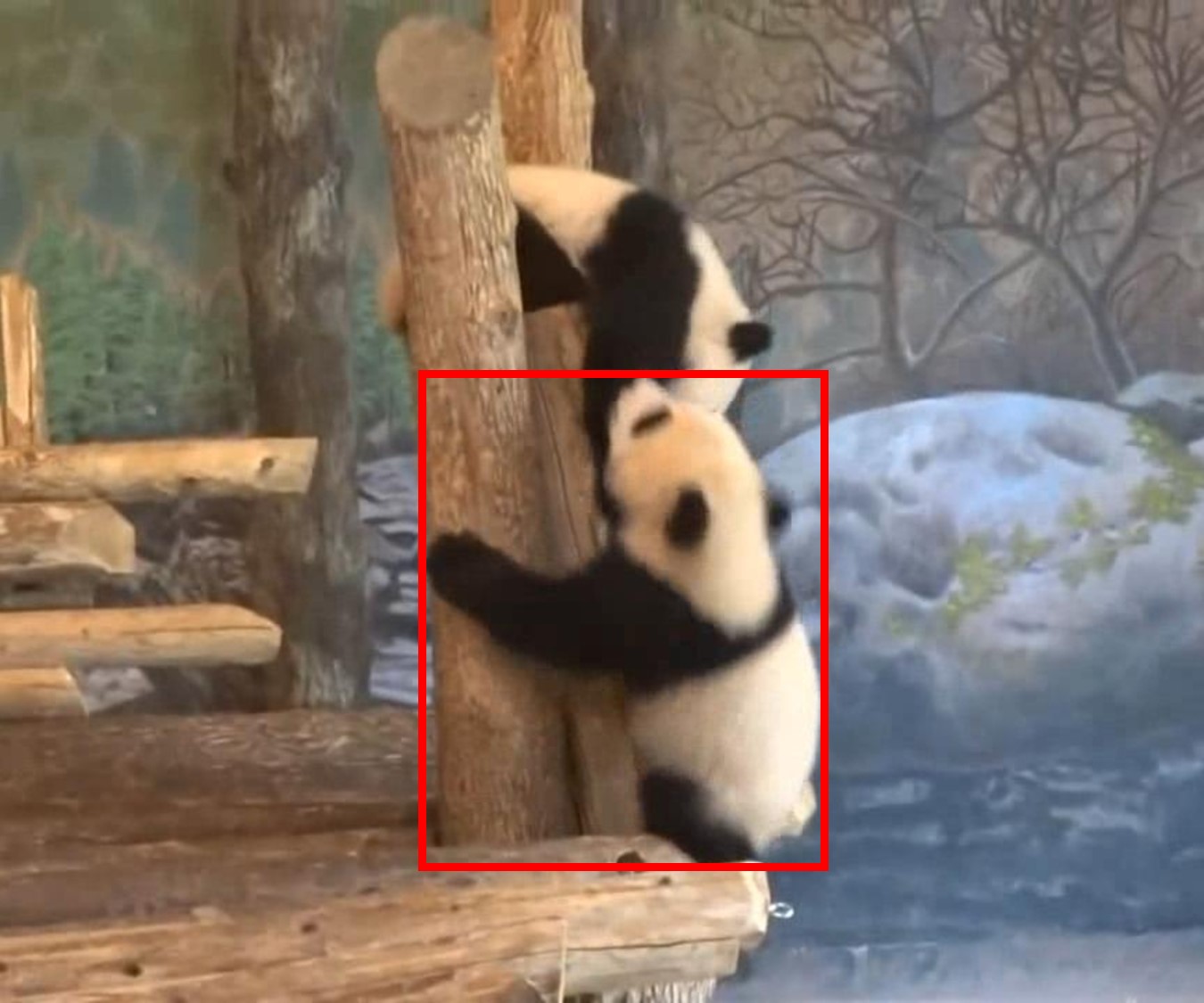}
    \includegraphics[width=\textwidth]{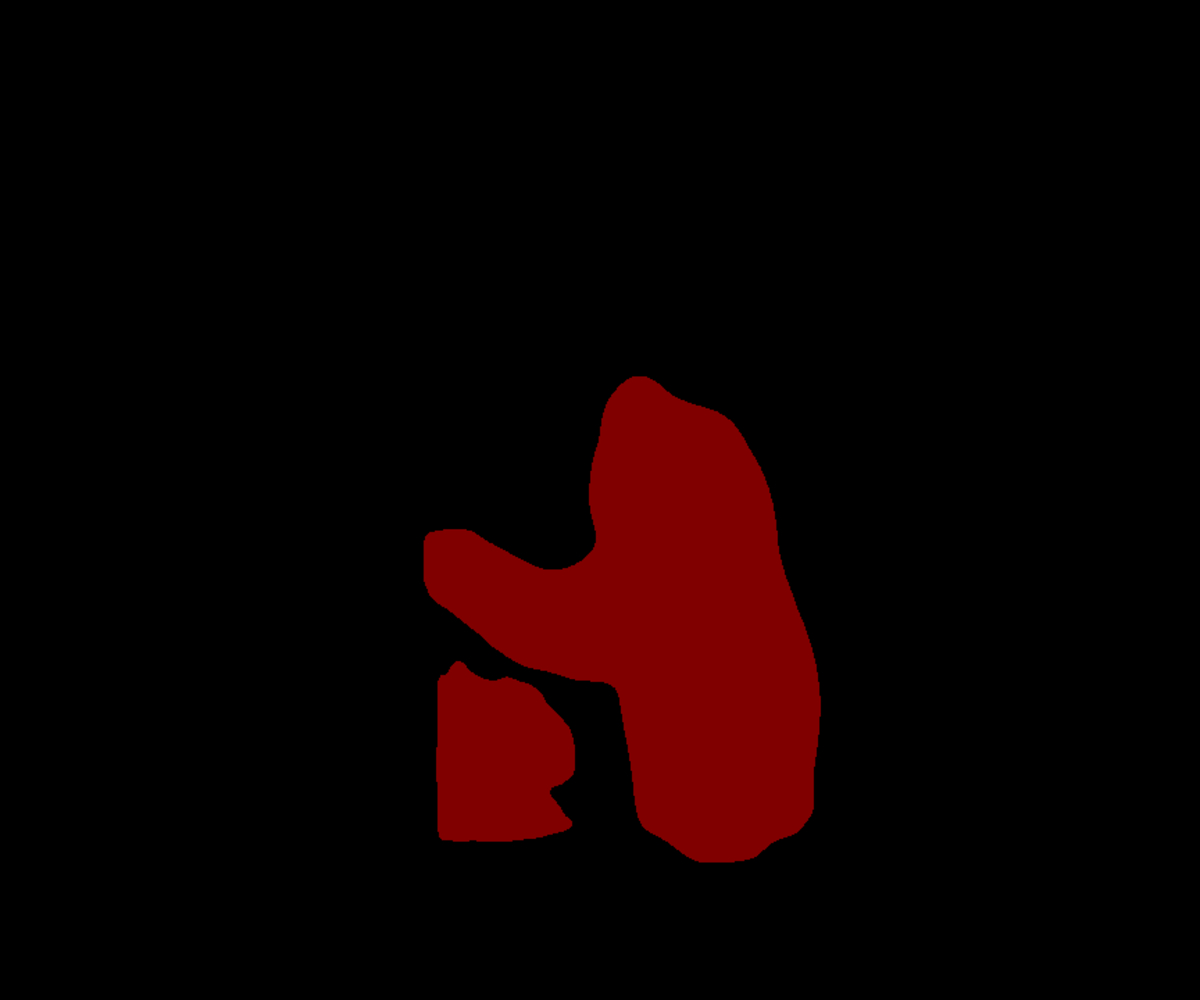}
    \includegraphics[width=\textwidth]{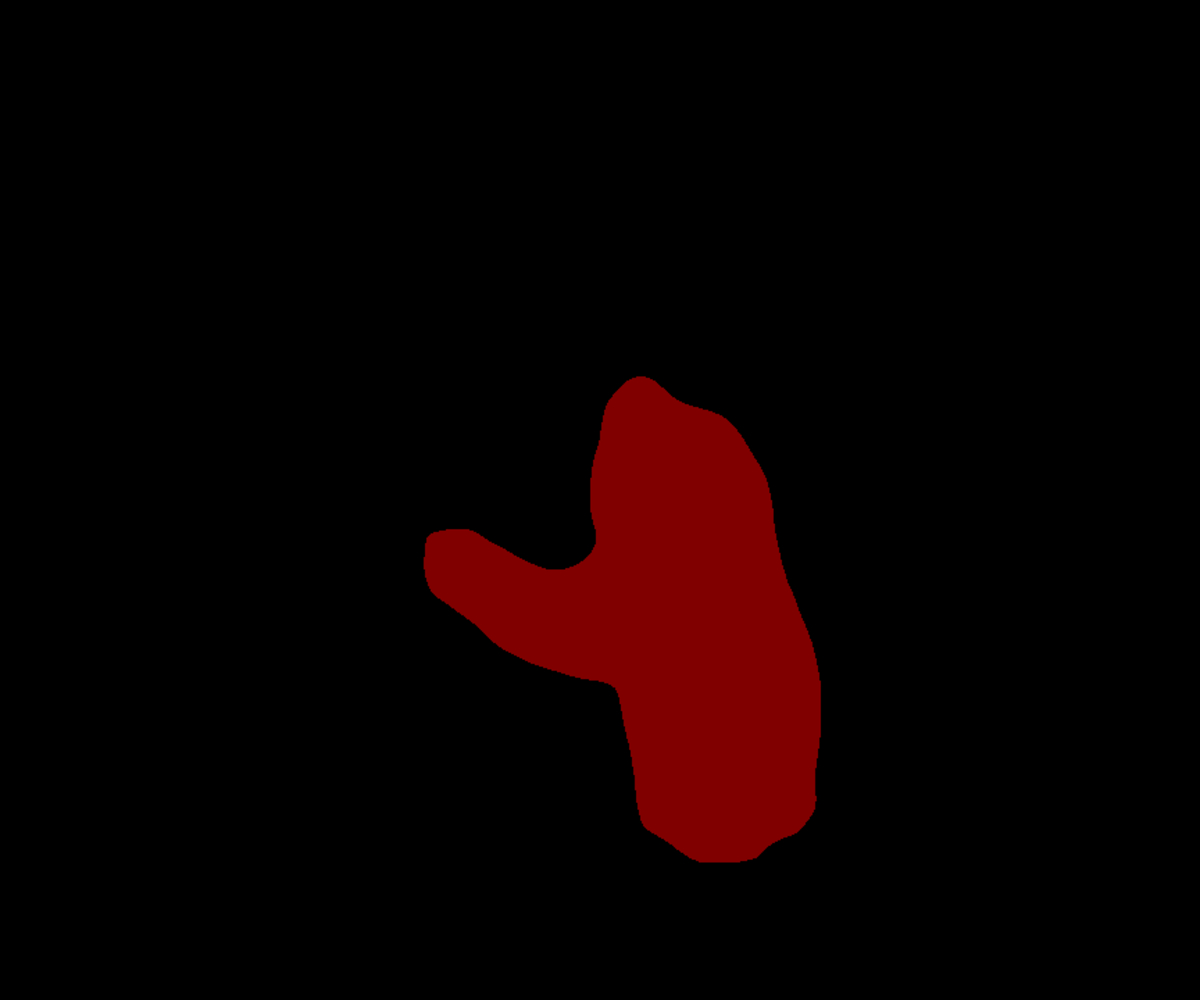}
	\end{minipage}
	}\hspace{-2mm}
	\subfloat{
    \begin{minipage}[c]{0.12\textwidth}
    \includegraphics[width=\textwidth]{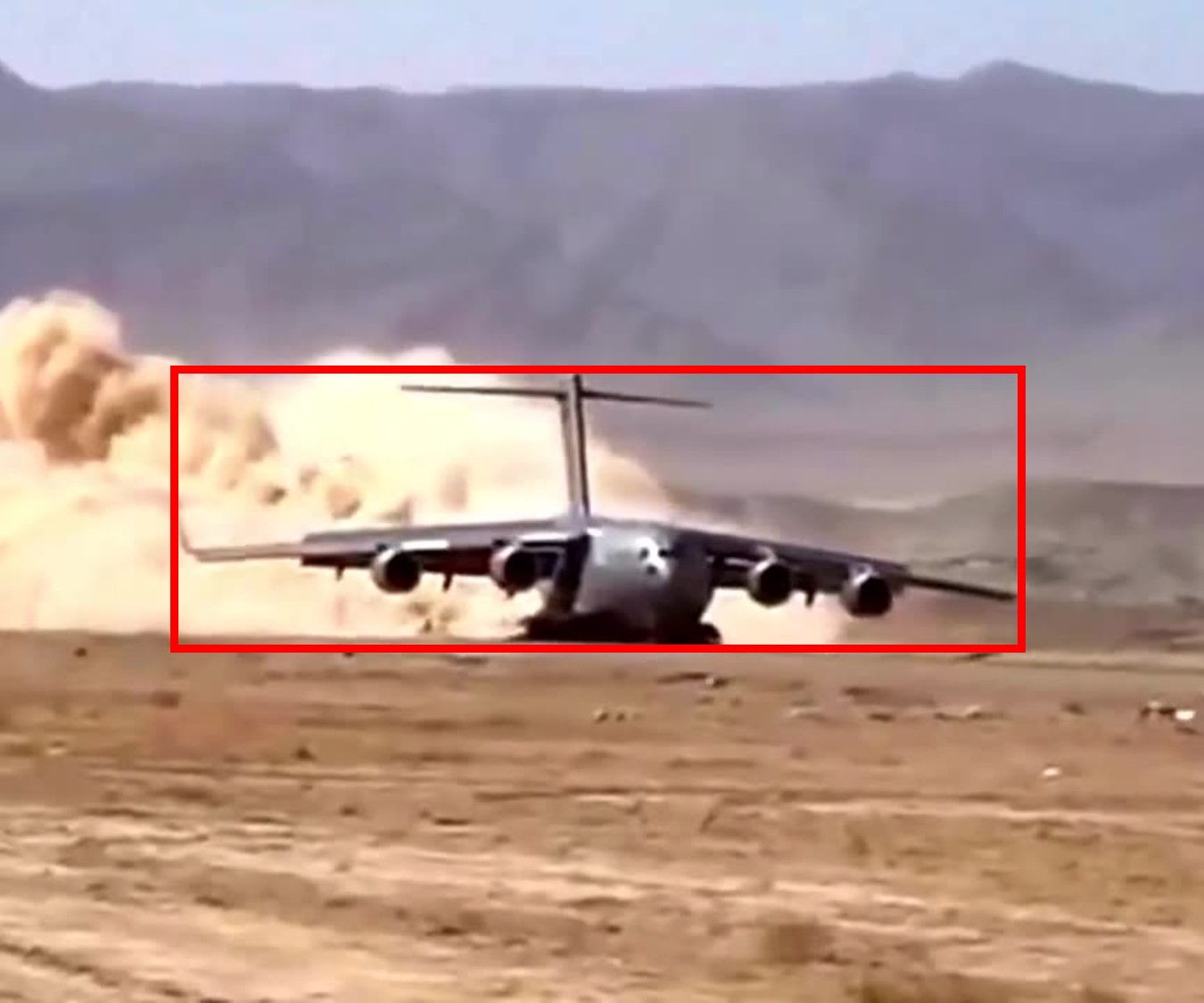}
    \includegraphics[width=\textwidth]{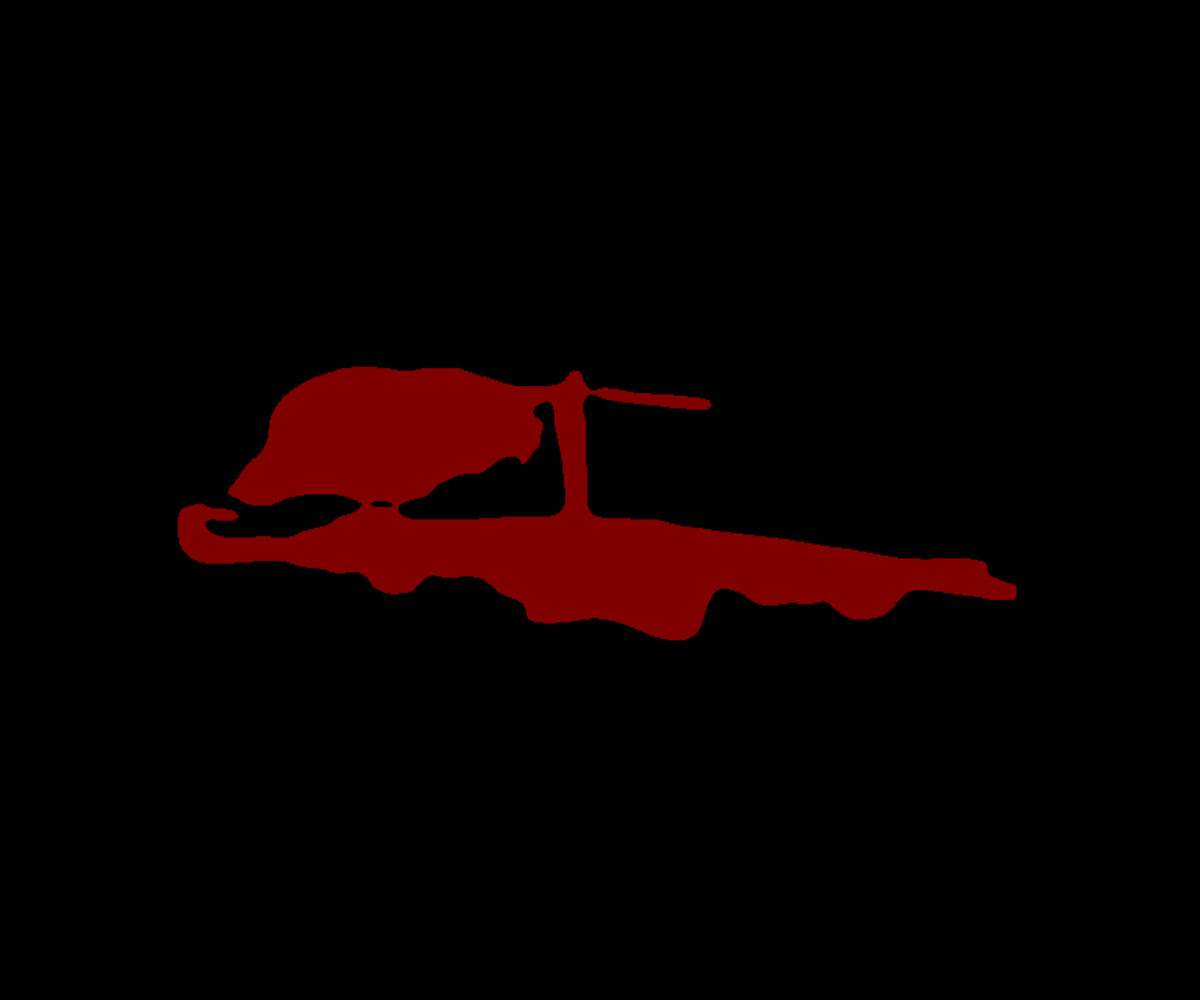}
    \includegraphics[width=\textwidth]{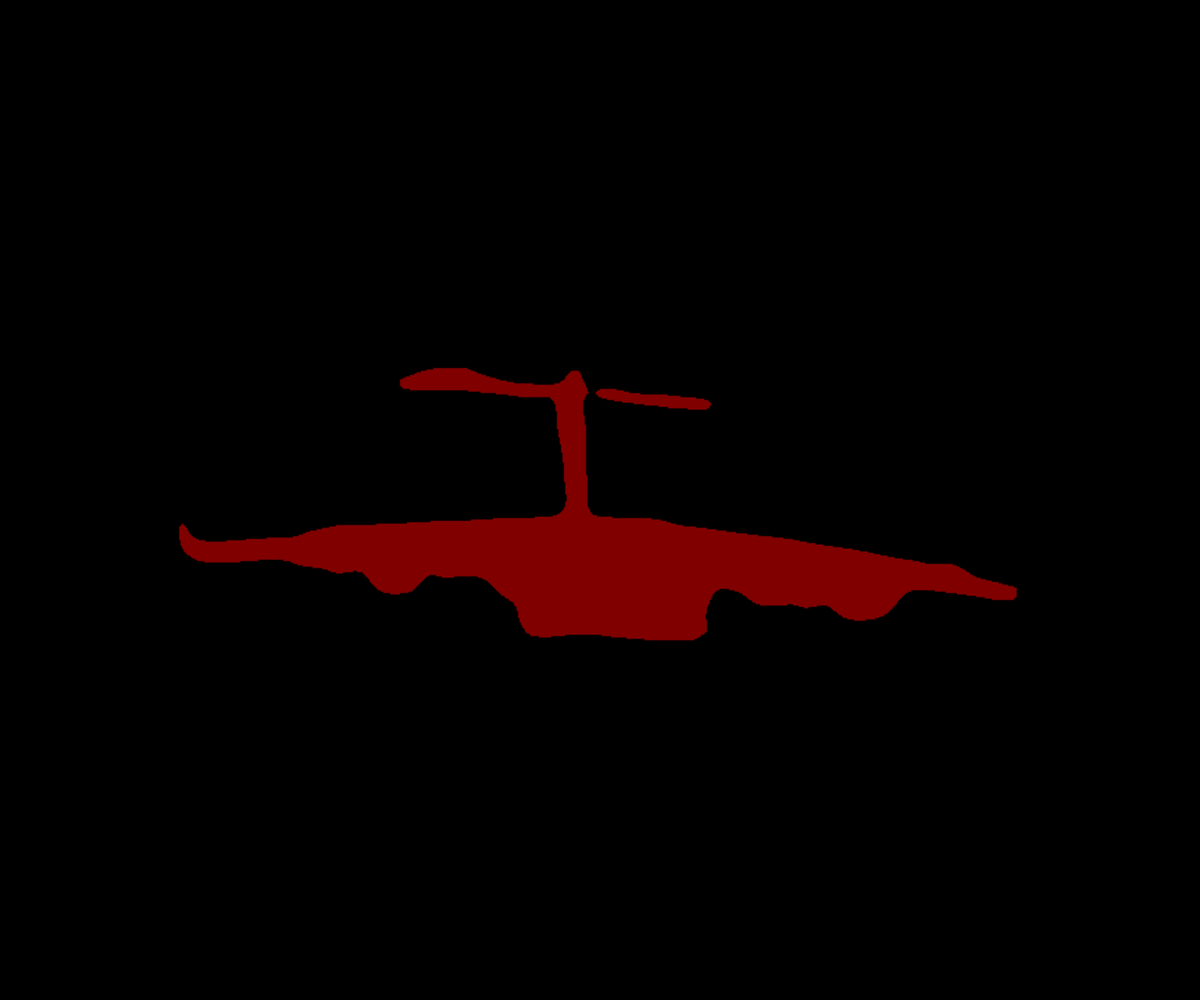}
	\end{minipage}
	}\hspace{-2mm}
	\subfloat{
    \begin{minipage}[c]{0.12\textwidth}
    \includegraphics[width=\textwidth]{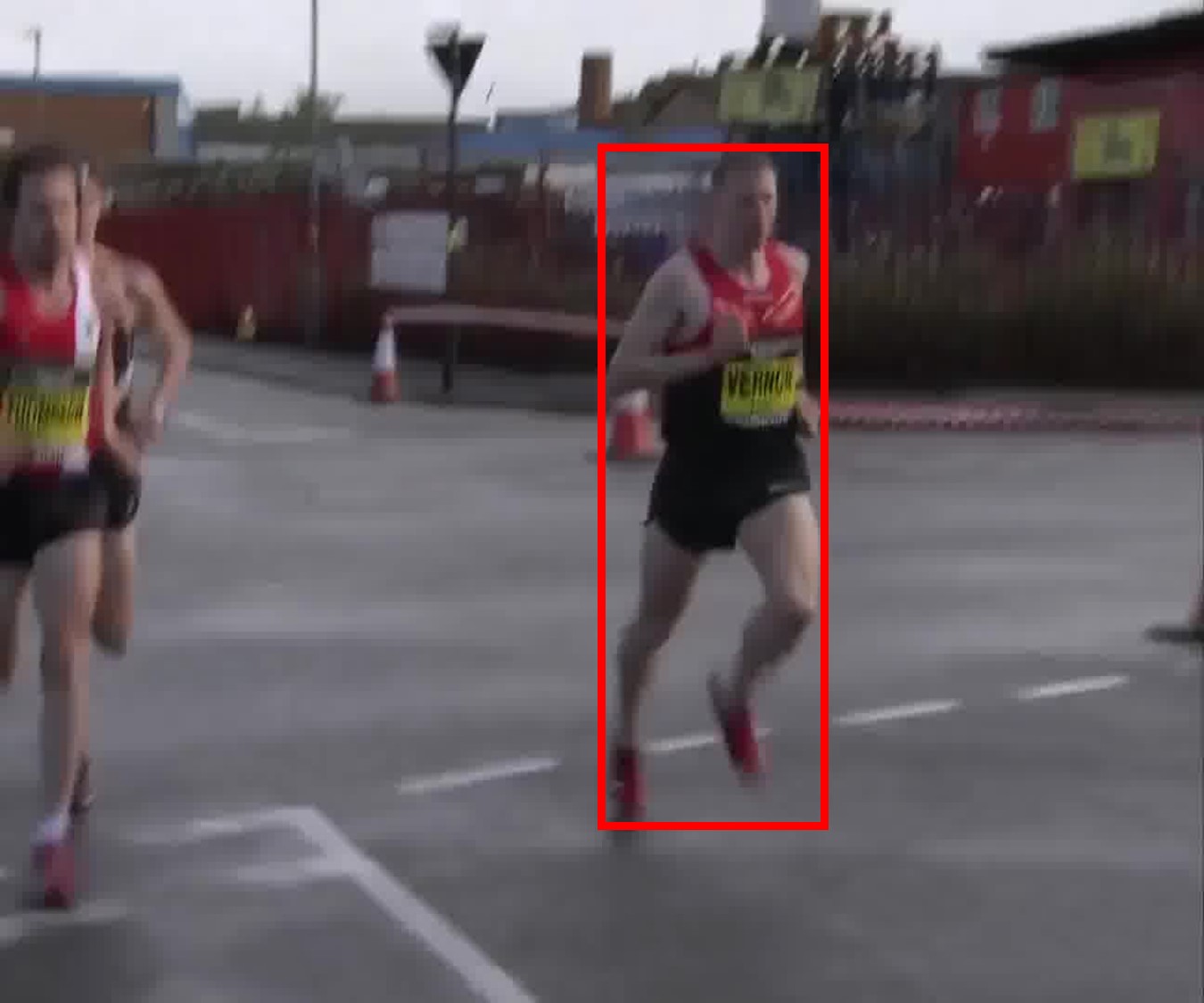}
    \includegraphics[width=\textwidth]{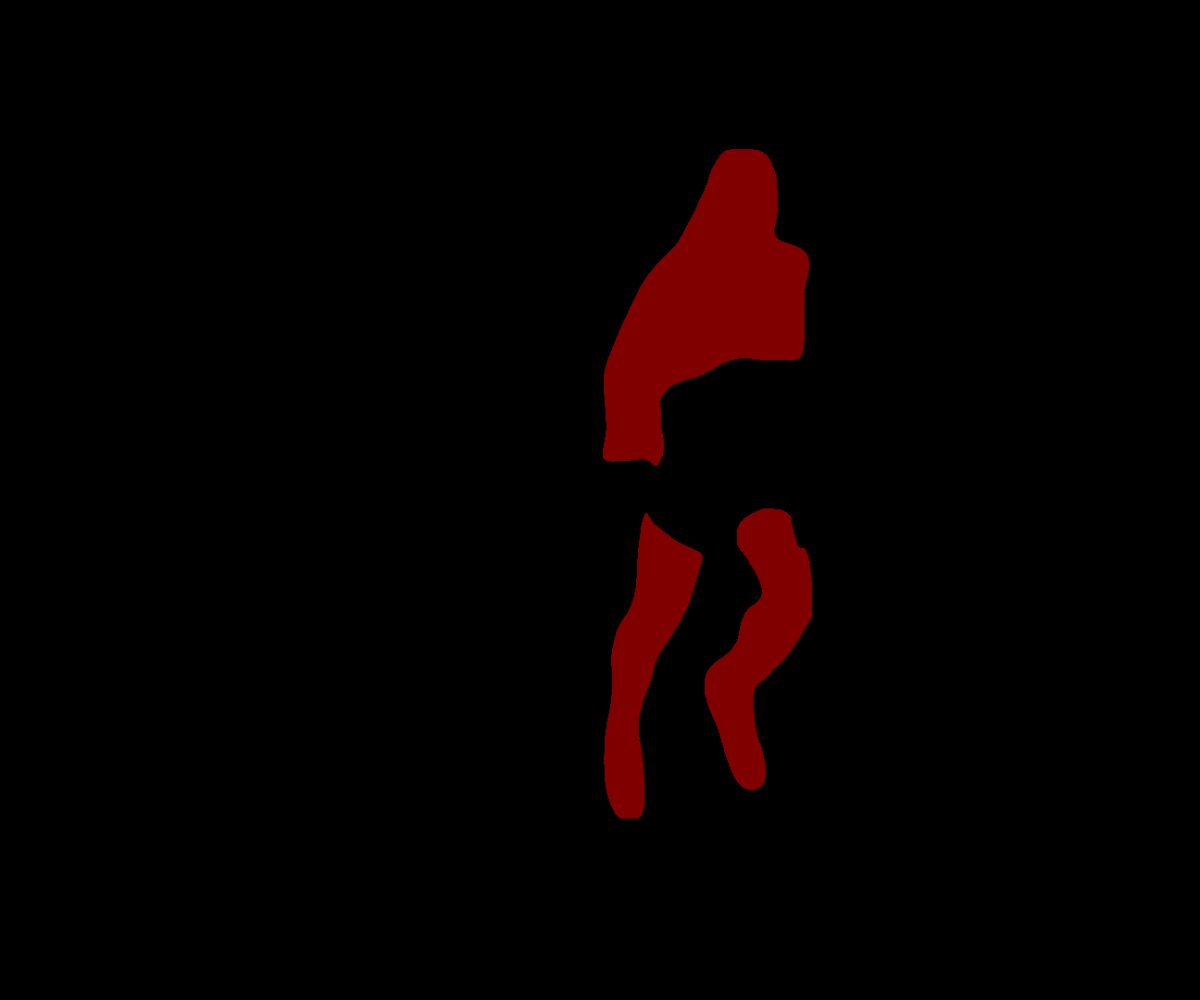}
    \includegraphics[width=\textwidth]{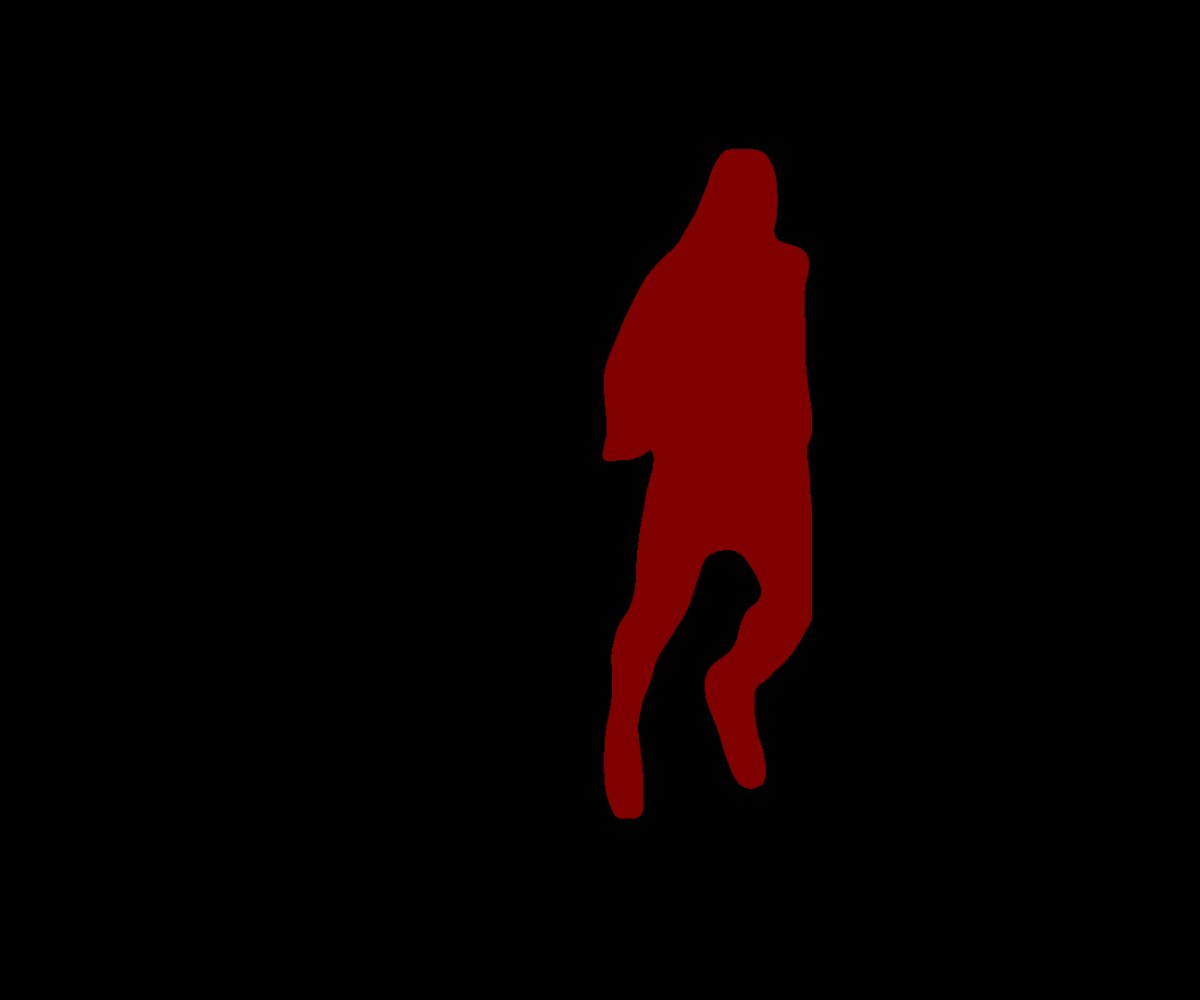}
	\end{minipage}
	}\hspace{-2mm}
	\subfloat{
    \begin{minipage}[c]{0.12\textwidth}
    \includegraphics[width=\textwidth]{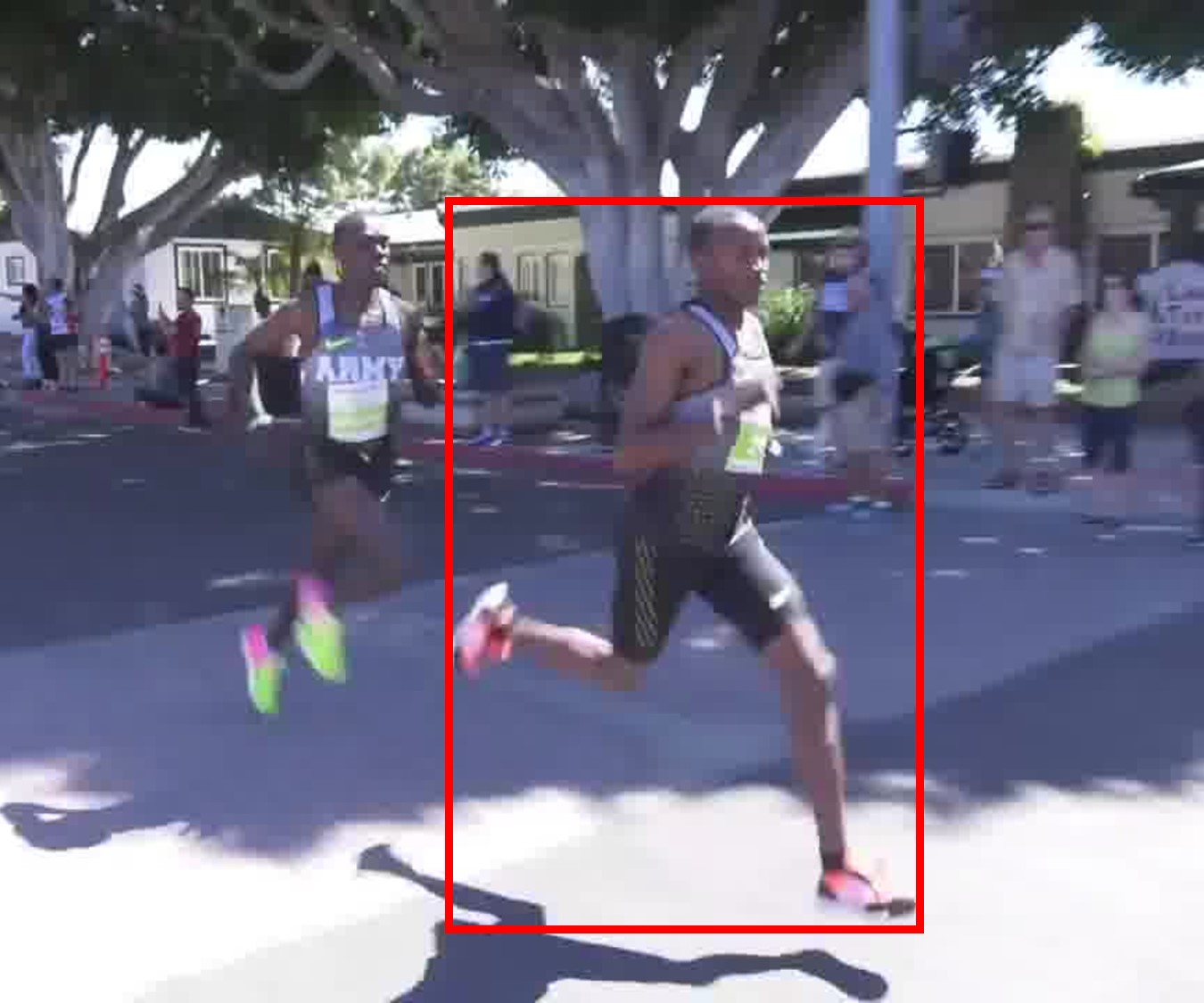}
    \includegraphics[width=\textwidth]{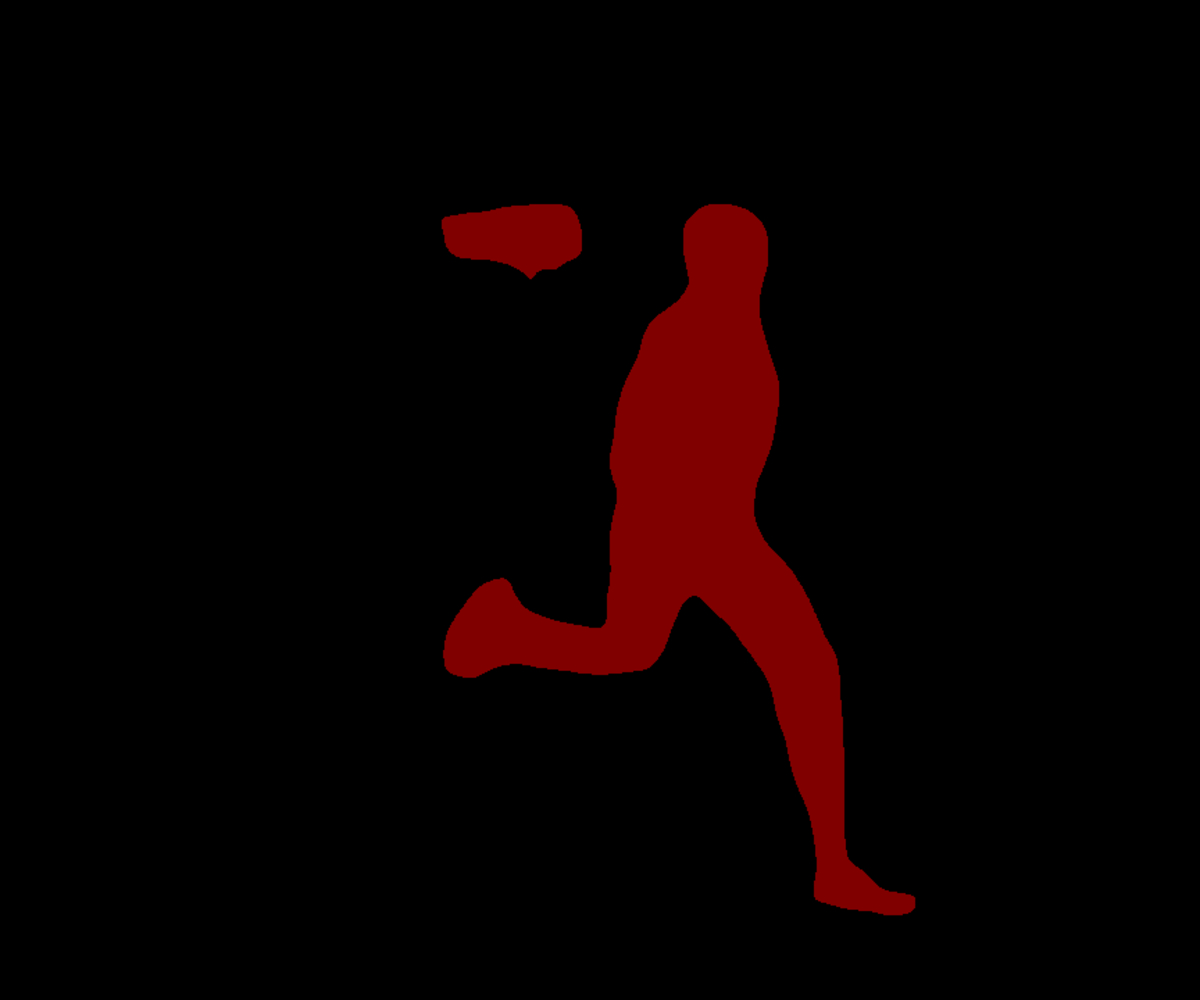}
    \includegraphics[width=\textwidth]{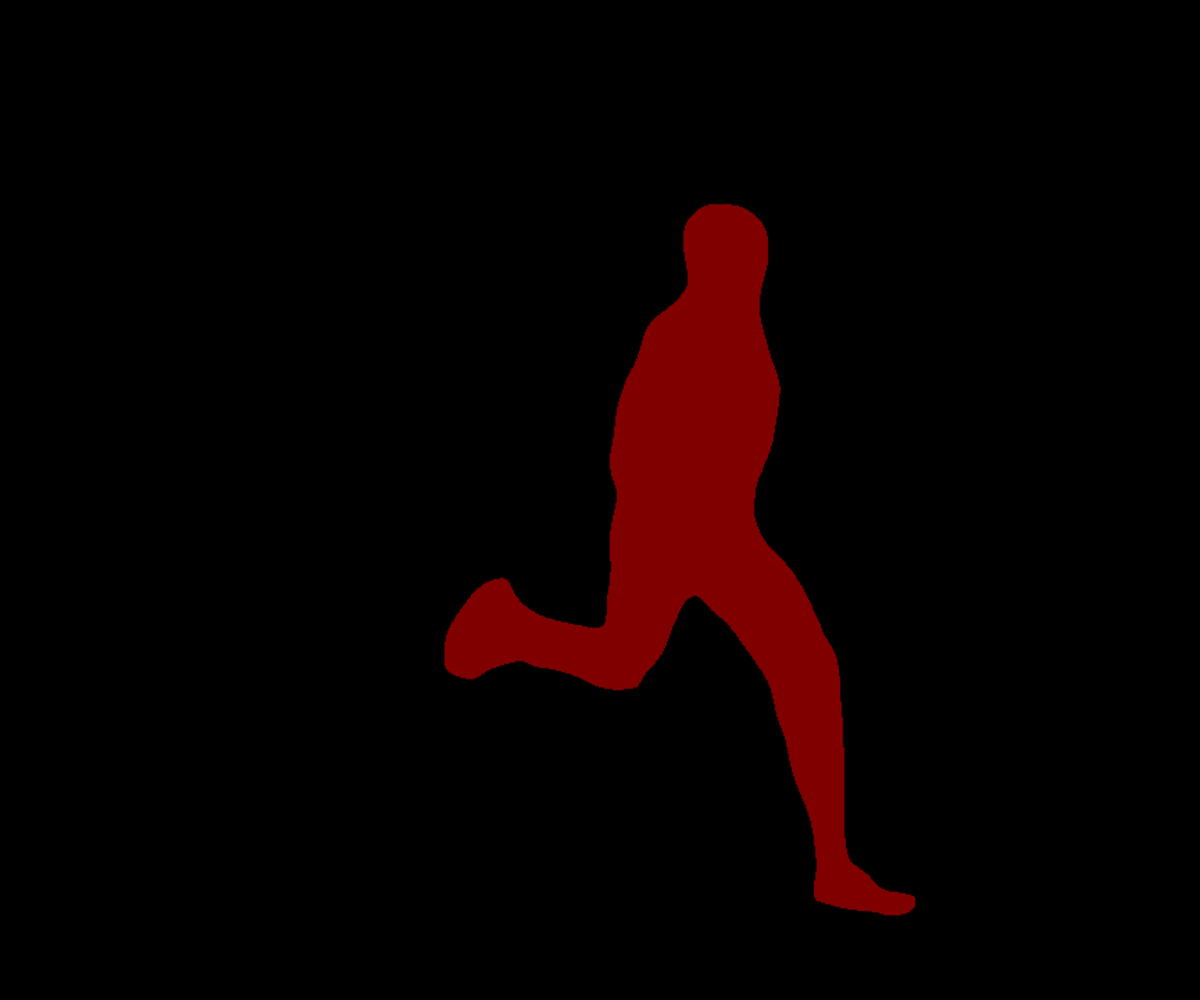}
	\end{minipage}
	}
    \caption{Qualitative results of our box to mask conversion network when using single (second row) and multiple (third row) images during inference. Our approach can effectively exploit multiple frames to handle challenging cases by mining spatio-temporal consistencies.}\vspace{-4mm}
    \label{fig:qual_fig}
\end{figure*}

\section{Experiments}
We perform comprehensive experiments to validate our contributions. A detailed ablative analysis of our architecture is provided in Sec.~\ref{sec:ablation}. We demonstrate the effectiveness of our approach for weakly supervised training of VOS in Sec~\ref{sec:wsvos}. Finally, in Sec.~\ref{sec:tracking_vos}, we use our network to annotate large-scale tracking datasets and show improved performance on tracking datasets using the generated annotations. 

\subsection{Ablation Study}
\label{sec:ablation}
Here, perform a detailed ablative study, analysing the impact of the key components in the proposed approach. Our analysis is performed on a custom validation set YT300 consisting of 300 sequences sampled randomly from the YouTube-VOS 2019 training set, as well as the DAVIS 2017 validation set. The methods are evaluated using the mean Jaccard $\mathcal{J}$ index. Unless specified, inference is performed using the settings described in Section~\ref{sec:implementation}. We only employ a different $\Delta = 5$ for DAVIS due to faster motions.

\parsection{Impact of using multiple frames} We investigate the impact of exploiting temporal information from multiple frames to convert boxes to masks by evaluating our approach using different number of input frames. Additionally, we also include an off-the-shelf single frame box to mask conversion network Box2Seg~\cite{luiten2018premvos} for comparison. Box2Seg uses a DeepLabv3+ architecture and directly predicts the mask using the image crop denoted by the box as input. The result of this comparison is shown in Table~\ref{tab:impact_of_frames}, and qualitative examples are provided in Fig. \ref{fig:qual_fig}. When using a single frame as input, our approach obtains a $\mathcal{J}$ score of $84.2$ and $78.7$ on YT300 and DAVIS 2017 validation set, respectively. The performance of our approach improves substantially when using multiple input frames. The best results are obtained when using $9$ frames. In this setting, our approach obtains a $\mathcal{J}$ score of $81.2$ on the DAVIS 2017 validation set, outperforming the single frame Box2Seg network by $+1.9$ in $\mathcal{J}$ score. These results clearly demonstrate the advantages of using multiple frames to perform accurate box to mask conversion.

\begin{table}[t]
    \centering
    \resizebox{\columnwidth}{!}{
    \begin{tabular}{l|c|cccccc}
    \toprule
         &Box2Seg&\multicolumn{6}{c}{Ours} \\\hline
         Num. Frames&1&1&3&5&7&9&11\\
         \midrule
         YT300&-&84.2&85.2&85.5&85.6&85.6&85.6\\
         DAVIS2017 val&79.3&78.7&80.4&80.9&81.1&81.2&81.2\\
    \bottomrule
    \end{tabular}
    }\vspace{2mm}
    \caption{Impact of using multiple frames for box to mask conversion. Results are shown in terms of Jaccard $\mathcal{J}$ index.}\vspace{-2mm}
    \label{tab:impact_of_frames}
\end{table}

\parsection{Analysis of architecture} Here, we analyse the impact of different components in our architecture. We evaluate four different variants of our method; i) \textbf{SingleImage:} A single image baseline which only uses the single-frame object representation $m_t$ to independently convert boxes to masks in each frame.  ii) \textbf{MultiFrame:} Our spatio-temporal aggregation module is used to obtain the segmentation encoding $s_t$ by exploiting multiple frames. 
iii) \textbf{MultiFrame+:} The single-frame object representation $m_t$ is passed to the segmentation decoder, in addition to the segmentation encoding $s_t$. iv) \textbf{MultiFrameIterative:} We employ the iterative refinement strategy described in Sec~\ref{sec:iterative-model} to refine the initial segmentation prediction obtained using \textbf{MultiFrame+}. Results are shown in Table~\ref{tab:ablation study ResNet50}. The SingleImage achieves a $\mathcal{J}$ score of $83.3$ and $77.2$ on YT300 and DAVIS 2017 validation set, respectively. The MultiFrame model, which exploits object information from multiple frames achieves significantly better results, with an improvement of $+2.6$ in $\mathcal{J}$ score on DAVIS 2017. This demonstrates the effectiveness of our spatio-temporal aggregation module in effectively combining the information from multiple frames. Using the single-frame object representation $m_t$ in combination with the segmentation encoding $s_t$  provides a slight improvement. Finally, performing iterative refinement of the initial segmentation prediction provides an additional improvement of $+1.1$ in $\mathcal{J}$ score on DAVIS 2017. This shows that the segmentation decoder contains rich prior information which can complement our spatio-temporal aggregation module.

\begin{table}[t]
    \centering
    \resizebox{\columnwidth}{!}{
    \begin{tabular}{l|ccc|cc}
    \bottomrule
    &$m_t$&$s_t$&Iter. &YT300&DAVIS2017 val\\
    \hline
    SingleImage& \checkmark & & & 83.3 & 77.2\\
    MultiFrame &&\checkmark& & 84.6 & 79.8\\
    MultiFrame+&\checkmark &\checkmark & & 84.8 & 80.1\\
    MultiFrameIterative&\checkmark &\checkmark &\checkmark & 85.6 & 81.2\\
      \bottomrule
    \end{tabular}
    }\vspace{1mm}
    \caption{Impact of different components in the proposed approach. Results are reported in terms of Jaccard $\mathcal{J}$ score.}\vspace{-5mm}
    \label{tab:ablation study ResNet50}
\end{table}

\subsection{Weakly Supervised VOS Training}
\label{sec:wsvos}
In this section, we validate the effectiveness of our approach to generate pseudo-labels for weakly supervised training of VOS models. We consider the scenario where pixel-wise segmentation labels are only available for a small number of training sequences, while the rest of the sequences have bounding box annotation for the objects. This is a highly practical scenario as generating bounding box labels is significantly faster compared to obtaining pixel-wise mask annotations. In such cases, it is desirable to exploit the bounding box annotations to perform weakly supervised training in order to benefit from more training data. 
In order to evaluate our approach for this weakly supervised setting, we simulate the training scenario using YouTube-VOS 2019 training set. We randomly split YouTube-VOS training set into two subsets A and B in the ratio 1:9. The segmentation labels are available for set A, while only the bounding box annotations are made available for videos in set B.

\begin{table}[t]
    \centering
    \resizebox{0.7\columnwidth}{!}{
    \begin{tabular}{l|ccccc}
    \toprule
    &Only A&MIL&MIL+CRF&\textbf{Ours}&FS \\
    \midrule
     $\mathcal{J}$&76.9&77.7&77.8&78.9&79.8 \\
     \bottomrule
    \end{tabular}
    }\vspace{1mm}
    \caption{Comparison with other weakly supervised training methods on the YT300 dataset, in terms of Jaccard $\mathcal{J}$ Index.}
    \vspace{-5mm}
    \label{tab:wvos}
\end{table}

We use the mask annotated videos from set A to train our video box to mask conversion network. The trained model is then used to generate pseudo-labels for video in set B, using only the bounding box annotation. A VOS model is then trained using the combined datasets A and B. We compare our approach of generating pseudo labels using video with two alternative; i) \textbf{MIL} We use the recently introduced multiple instance learning (MIL) loss \cite{hsu2019weakly} to compute training loss on the box annotated videos from set B. ii) \textbf{MIL+CRF} We use the MIL loss in combination with the CRF regularizer introduced in \cite{tang2018regularized} to compute training loss. Additionally, we also report the results obtained when using only the fully annotated set A for training (\textbf{Only A}), as well as the upper bound attained when using the complete YouTube VOS training set with mask annotations (\textbf{FS}). We use the recently introduced LWL~\cite{bhat2020learning} approach as our VOS model for this experiment. The LWL network is trained using each of the weakly supervised methods, for 100k iterations. The result of this comparison is shown in Table~\ref{tab:wvos}, on the YT300 set. Both the \textbf{MIL} and \textbf{MIL+CRF} approaches obtain an improvement of around $+0.9$ in $\mathcal{J}$ score, compared to the nai\"ve baseline using only the mask annotated videos from set A for training. Our approach of generating pseudo labels  obtains the best results, achieving a substantial improvement of over $+1$ in $\mathcal{J}$ score over the MIL baselines. These results demonstrate the quality and effectiveness of the masks generated from our approach for performing weakly supervised training for VOS. 

\begin{table}[t]
    \centering
    \scriptsize
    \setlength{\tabcolsep}{0.75mm}
    \begin{tabular}{lcccccc}
    \toprule
     & STM& AlphaRef & OceanPlus & RPT & LWL & \textbf{LWL-Ours}\\
        \midrule
    EAO & 0.308&  0.482 & 0.491 & \textcolor{red}{0.530} & 0.463 & \textcolor{blue}{0.510}\\
    Accuracy & \textcolor{blue}{0.751} & \textcolor{red}{0.754} & 0.685 &0.700 & 0.719 & 0.732\\
    Robustness & 0.574 & 0.777 & \textcolor{blue}{0.842} & \textcolor{red}{0.869} & 0.798 & 0.824\\
    \bottomrule
    \end{tabular}\vspace{1mm}
    \caption{State-of-the-art comparison on VOT2020 in terms of expected average overlap (EAO), accuracy, and robustness.}\vspace{-3mm}
    \label{tab:vot}
\end{table}

\begin{table}[]
    \centering
    \scriptsize
    \setlength{\tabcolsep}{0.75mm}
    \begin{tabular}{lccccc}
    \toprule
         & SiamRPN++ \cite{li2019siamrpn++} & DiMP-50 \cite{bhat2019learning} & PrDiMP-50 \cite{danelljan2020probabilistic} & LWL & \textbf{LWL-Ours}\\
         \midrule
    $\text{SR}_{0.5}$ (\%) & 82.8 & 88.7 & 89.6 & \textcolor{blue}{92.4} & \textcolor{red}{95.1}\\
    $\text{SR}_{0.75}$ (\%) & - & 68.8 & 72.8 & \textcolor{blue}{82.2} & \textcolor{red}{85.2}\\
    AO (\%) & 73.0 &  75.3 & 77.8 & \textcolor{blue}{84.6} & \textcolor{red}{86.7}\\
    \bottomrule
    \end{tabular}\vspace{1mm}
    \caption{State-of-the-art comparison on the GOT-10k validation set in terms of average overlap (AO) and success rates (SR) at overlap thresholds 0.5 and 0.75}\vspace{-3mm}
    \label{tab:got10k-val}
\end{table}

\subsection{VOS in the Tracking Domain}
\label{sec:tracking_vos}
We utilize the capability of performing weakly supervised VOS training using box annotations to train a VOS method on large-scale tracking datasets in order to obtain improved tracking performance. We use our network to annotate large-scale tracking datasets LaSOT~\cite{fan2019lasot} and GOT10k \cite{Huang2019} containing 1120 and 9340 training sequences, respectively. The pseudo annotated tracking sequences, along with the fully annotated YouTube-VOS and DAVIS datasets are then used to fine-tune a VOS model. We start with the LWL~\cite{bhat2020learning} network trained with fixed backbone weights. The complete model, including the backbone feature extractor, is then trained on combined YouTube-VOS, DAVIS, LaSOT, and GOT-10k datasets for 120k iterations. We compare this model, denoted \textbf{LWL-Ours}, with the state-of-the-art on VOT2020 \cite{kristannew}, GOT10K \cite{Huang2019}, and TrackingNet \cite{muller2018trackingnet} datasets. For comparison, we also report results for the standard LWL model fine-tuned using only the YouTube-VOS and DAVIS datasets. 

\parsection{VOT2020 \cite{kristannew}} We evaluate our trained LWL-Ours model on VOT2020 dataset consisting of 60 challenging sequences. Similar to semi-supervised VOS, the trackers are provided an initial object mask. 
In order to obtain robust performance measures, the trackers are evaluated multiple times on each sequence, using different starting frames. The trackers are compared using the accuracy, robustness, and expected average overlap EAO measure. Accuracy denotes the average overlap between tracker prediction and the ground truth over the successfully tracked frames, while robustness measures the fraction of sequence tracked on average before tracking loss. Both these measures are combined to obtain the EAO score. Tab.\ \ref{tab:vot} shows the results over the 60 sequences from VOT2020. LWL-Ours fine-tuned on tracking datasets, obtains a relative improvement of over $10\%$ in EAO score, compared to the LWL baseline. Furthermore, despite performing vanilla VOS, LWL-Ours outperforms existing tracking approaches, achieving the second best EAO score. These results show that the masks generated from our approach can be utilized to improve the generalization of VOS model on generic tracking datasets.

\parsection{GOT10k \cite{Huang2019}} We evaluate LWL-Ours on the validation split of GOT10k dataset, consisting of 180 videos. Unlike VOT2020, trackers are only provided an initial box and required to output a target box for each frame. Thus, we use our box to mask conversion network to obtain the initial segmentation mask. The VOS model is then run using the generated mask. In each subsequent frame, we simply compute the target box using the extreme points of the predicted segmentation mask, without performing any post-processing. 
Fine-tuning LWL on our pseudo-annotated tracking videos provides an improvement of $2.1\%$ in AO over the baseline LWL model (see Tab.\ \ref{tab:got10k-val}). Moreover, LWL-ours significantly outperforms existing trackers with an AO score of $86.7\%$.

\parsection{TrackingNet \cite{muller2018trackingnet}} We report results on the test split of TrackingNet dataset consisting of 511 videos, using the same evaluation strategy employed for GOT10k dataset.
Using our generated masks on tracking datasets for fine-tuning improves the results of the LWL model by $0.5\%$ in terms of success score (see Tab. \ref{tab:trackingnet}). Moreover, LWL-Ours obtains the best results among all methods in terms of Success score, along with SiamRCNN~\cite{voigtlaender2020siam}. 

\begin{table}[]
    \centering
    \scriptsize
    \setlength{\tabcolsep}{0.75mm}
    \begin{tabular}{lcccccc}
    \bottomrule
     & \tabincell{c}{SiamRPN++ \\\cite{li2019siamrpn++}}& \tabincell{c}{DiMP-50 \\\cite{bhat2019learning}} & \tabincell{c}{KYS \\ \cite{bhat2020know}} & \tabincell{c}{SiamRCNN \\ \cite{voigtlaender2020siam}} & LWL & \textbf{LWL-Ours}\\
        \hline
    Precision (\%) & 69.4 & 68.7 & 68.8 &  \textcolor{red}{80.0} & 78.4 & \textcolor{blue}{79.1}\\
    Norm. Prec. (\%) & 80.0 & 80.1 & 80.0 & \textcolor{red}{85.4} & 84.4 & \textcolor{blue}{84.7}\\
    Success (AUC) (\%) & 73.3 & 74.0 & 74.0 & \textcolor{red}{81.2} & \textcolor{blue}{80.7} & \textcolor{red}{81.2}\\
    \bottomrule
    \end{tabular}\vspace{1mm}
    \caption{State-of-the-art comparison on the TrackingNet test set in terms of precision, normalized precision, and success.}\vspace{-3mm}
    \label{tab:trackingnet}
\end{table}

\begin{table}[]
    \centering
    \resizebox{0.7\columnwidth}{!}{
    \begin{tabular}{lccc}
    \toprule
         &  STM \cite{oh2019video} & LWL & \textbf{LWL-Ours}\\
         \midrule
     $\mathcal{J}\&\mathcal{F}$ mean & 79.2 & 81.0  & 80.6\\
    \bottomrule
    \end{tabular}
    }\vspace{1mm}
    \caption{Comparison on YouTube-VOS 2019 validation set.}
    \label{tab:ytvos19}\vspace{-5mm}
\end{table}

\parsection{YouTube-VOS} We also report results on the YouTube-VOS 2019 validation set for comparison. LWL-Ours obtains results comparable with standard LWL (see Tab. \ref{tab:ytvos19}), showing that the improved performance on tracking datasets is obtained without sacrificing VOS accuracy.

\section{Conclusion}
We propose an end-to-end trainable method for predicting object masks from bounding boxes in videos. Our approach can effectively mine object and background information over multiple frames using a novel spatio-temporal aggregation module. The predicted masks are further refined using an iterative formulation. Our approach obtains superior segmentation accuracy when converting boxes to mask in videos, compared to single image baselines. We further demonstrate the usefulness of our method for weakly supervised VOS training in limited data domain.

\parsection{Acknowledgements}
This work was partly supported by the ETH Z\"urich Fund (OK), a Huawei Gift for research, a Huawei Technologies Oy (Finland) project, an Amazon AWS grant, and an Nvidia hardware grant.

{\small
\bibliographystyle{ieee_fullname}
\bibliography{egbib}
}

\appendix
\newpage
\begin{center}
\Large\textbf{Appendix}
\end{center}
In this Appendix, we provide additional details and analysis of our approach. In Appendix~\ref{sec:obj_enc}, we provide details about the network architecture of our object encoder module. Appendix~\ref{sec:analysis} provides additional analysis on the impact of different hyper-parameters in the proposed approach. Details about the inference setting used to annotate the tracking datasets in Section 4.3 of the main paper are provided in Appendix~\ref{sec:inf_details}. Detailed results on the GOT10k validation set and YouTube-VOS 2018 validation set are provided in Appendix~\ref{sec:detailed_results}. Appendix~\ref{sec:qual} includes additional qualitative results on the DAVIS and GOT10k datasets. Additionally, we also include a video showing the results of our approach on the sequences from the DAVIS and GOT10k datasets.

\section{Included video}
We provide a video showing the masks produced by our approach for sequences from DAVIS and GOT10k datasets. Our approach generates high-quality masks even under difficult circumstances, such as fast motions, appearance change and shape change. The last sequence, \emph{bike-packing} from DAVIS, shows a particularly challenging case where two objects are highly overlapping.

\section{Structure of the object encoder}
\label{sec:obj_enc}
\begin{figure}[t]
    \centering
    \includegraphics[width=\columnwidth]{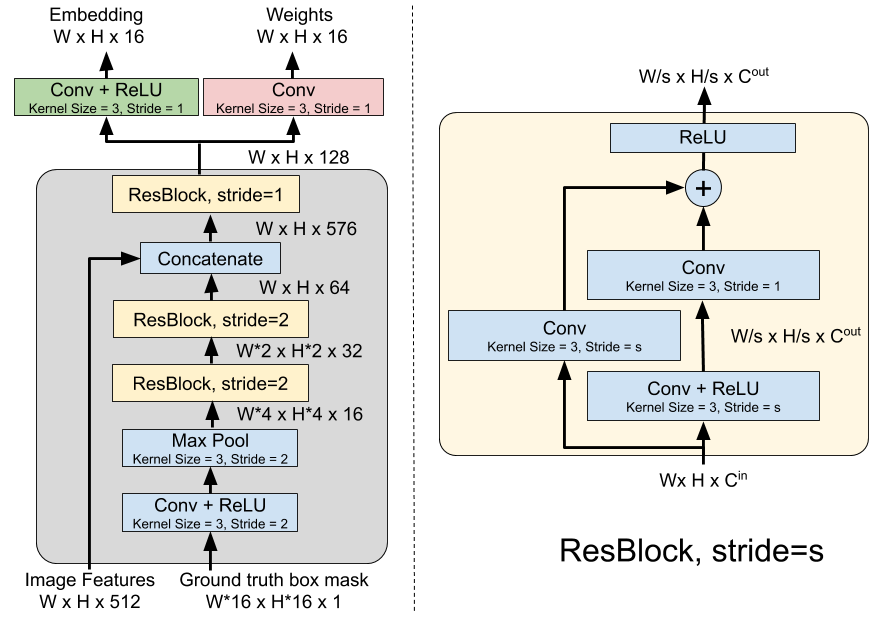}
    \caption{Architecture of object encoder.}
    \label{fig:object_encoder}
\end{figure}

Here, we describe in detail the network structure employed for the object encoder, as shown in Fig.~\ref{fig:object_encoder}. Note that our object encoder is formulated as $(e_t, w_t, m_t)=B(x_t, b_t)$. The network first takes as input the mask representation of the bounding box $b_t$ and then passes it to a convolution layer, a max pooling layer and two residual blocks. The intermediate mask features are then concatenated with deep features $x_t$ and fed through another residual block, which reduces the feature dimension. Finally, two similar heads are utilized to produce abstract embedding $e_t$ and weight $w_t$. Although the embedding and weight heads share common network to extract object representations, there is no need to share them for the single-frame encoding $m_t$. Single-frame encoding $m_t$ is directly passed to the decoder and has no connection with embedding $e_t$. Thus, we use a different network to obtain the single-frame encoding $m_t$. This network has the same architecture as the network used to obtain $e_t$ and $w_t$, with the only difference that it has a single head with a convolution and ReLU layers to predict $e_t$.

\begin{figure*}
    \centering
    \subfloat{
    \begin{minipage}[c]{0.195\textwidth}
    \includegraphics[width=\textwidth]{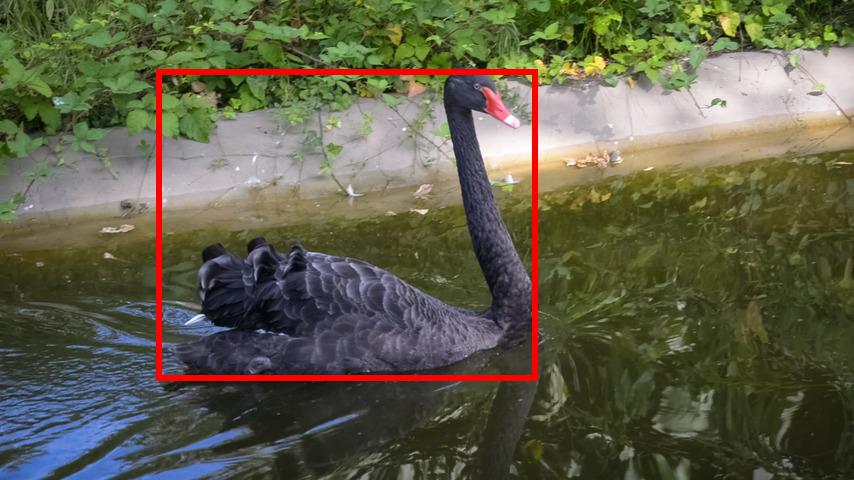}
    \includegraphics[width=\textwidth]{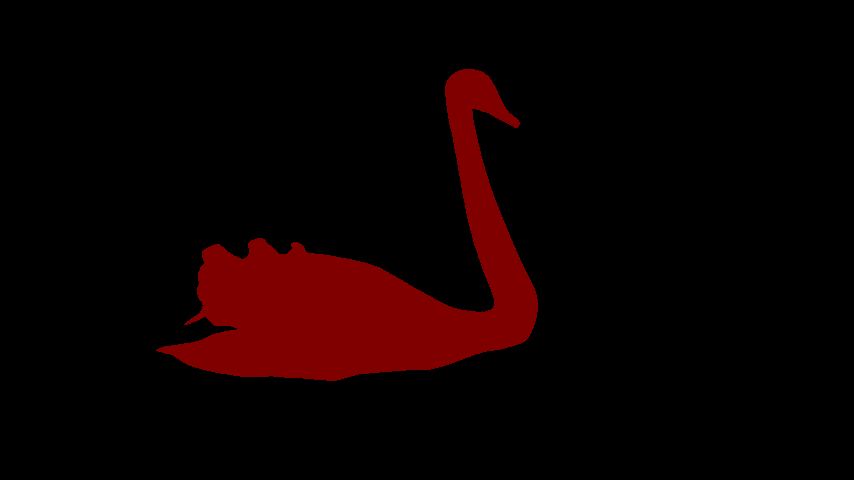}
    \includegraphics[width=\textwidth]{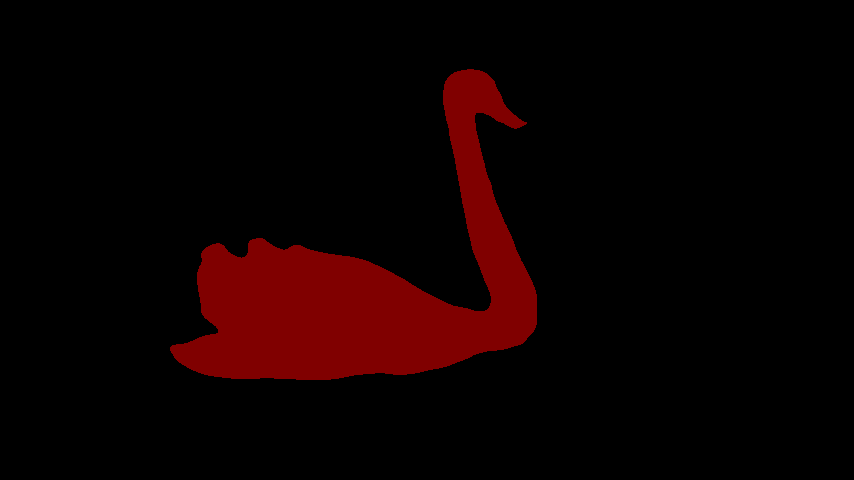}
	\end{minipage}
	}\hspace{-2mm}
	\subfloat{
    \begin{minipage}[c]{0.195\textwidth}
    \includegraphics[width=\textwidth]{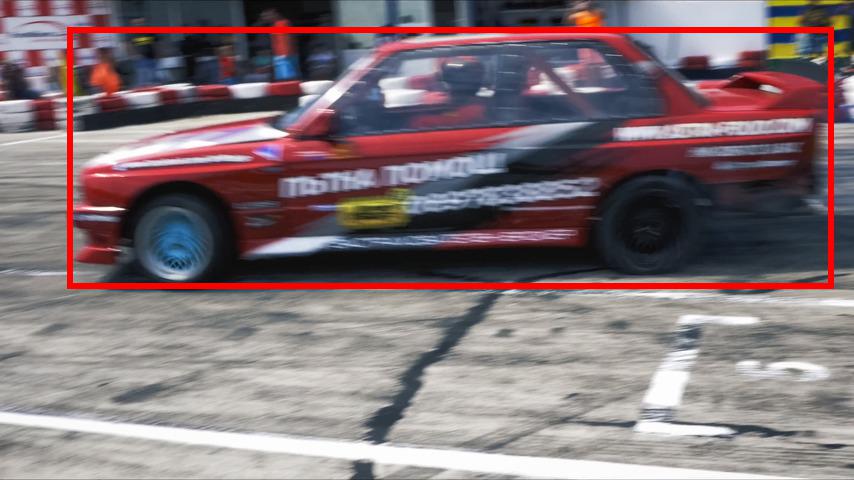}
    \includegraphics[width=\textwidth]{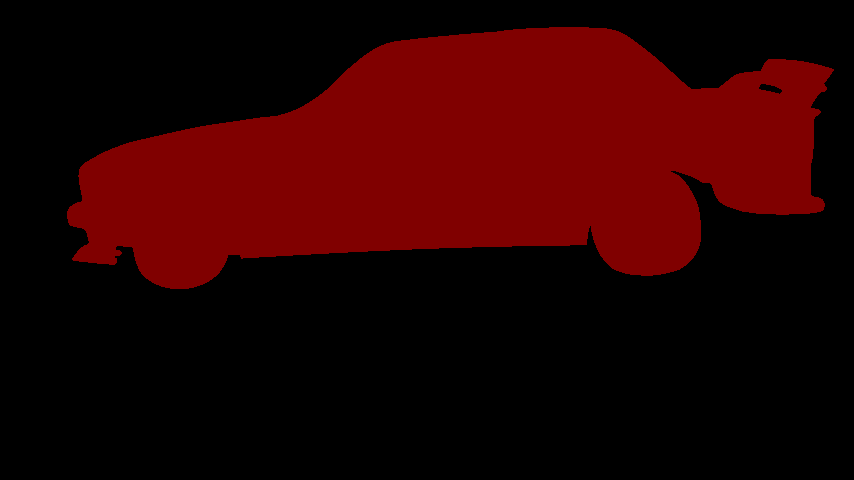}
    \includegraphics[width=\textwidth]{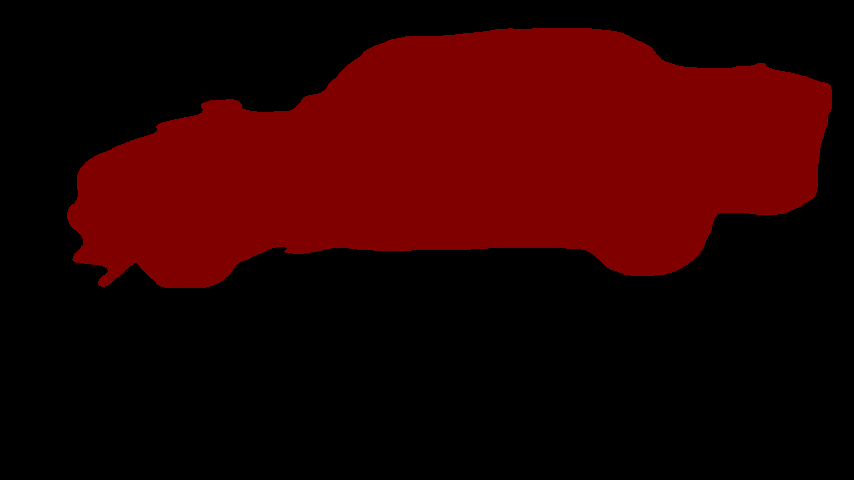}
	\end{minipage}
	}\hspace{-2mm}
	\subfloat{
    \begin{minipage}[c]{0.195\textwidth}
     \includegraphics[width=\textwidth]{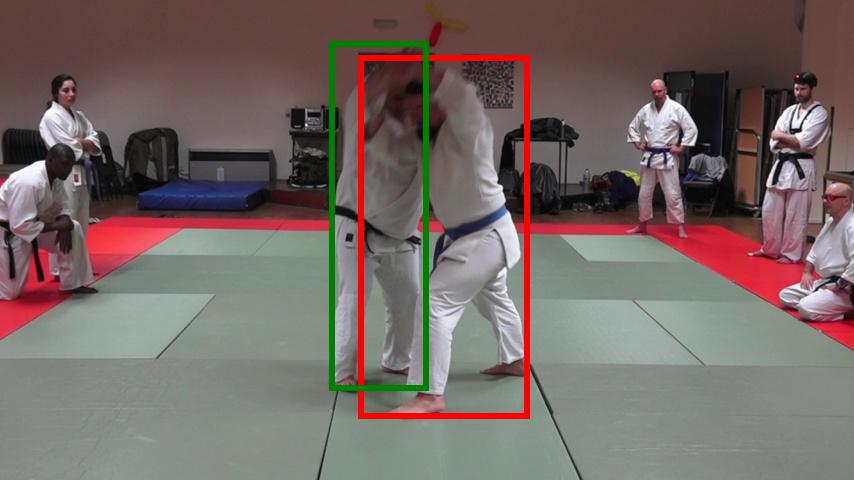}
     \includegraphics[width=\textwidth]{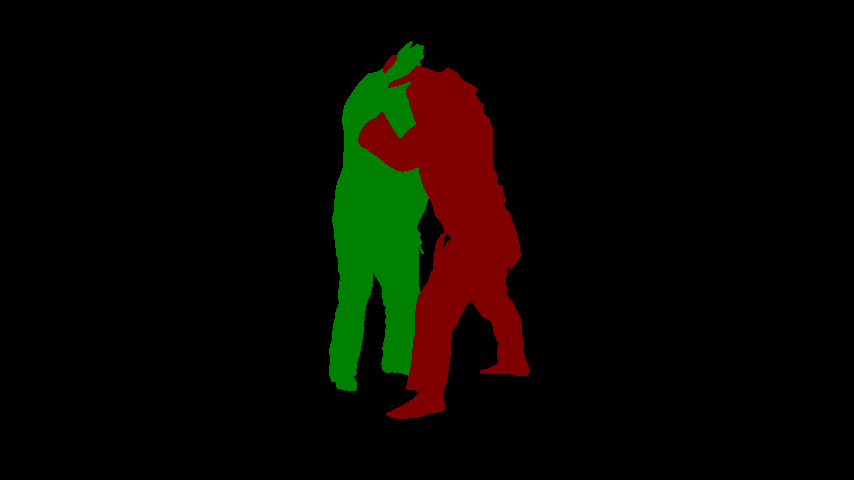}
     \includegraphics[width=\textwidth]{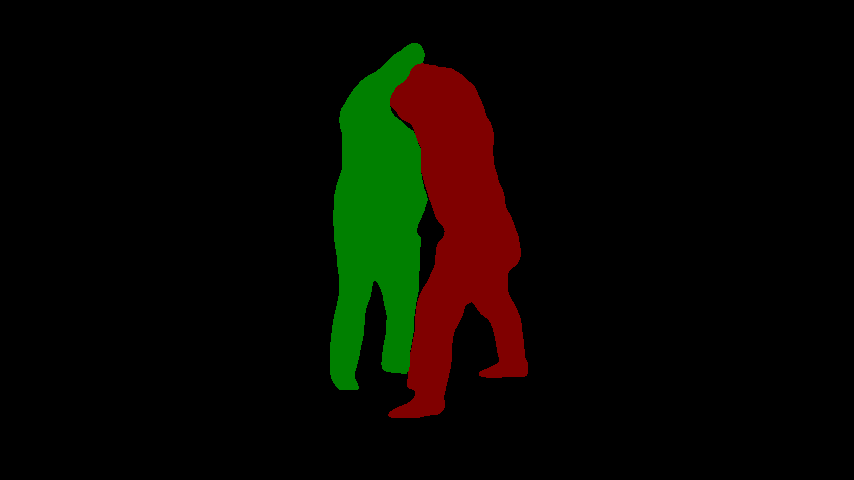}
	\end{minipage}
	}\hspace{-2mm}
	\subfloat{
	\begin{minipage}[c]{0.195\textwidth}
	\includegraphics[width=\textwidth]{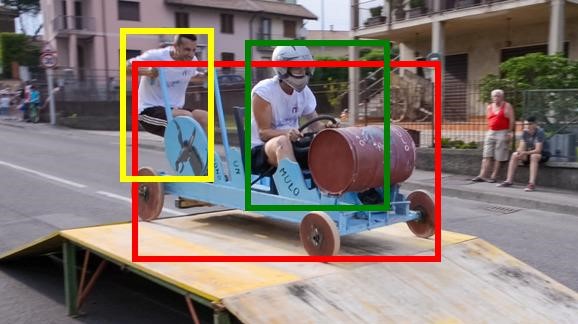}
    \includegraphics[width=\textwidth]{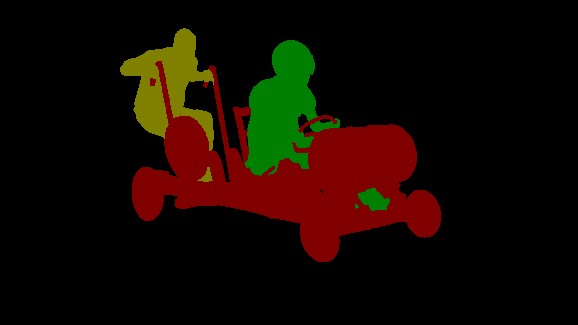}
    \includegraphics[width=\textwidth]{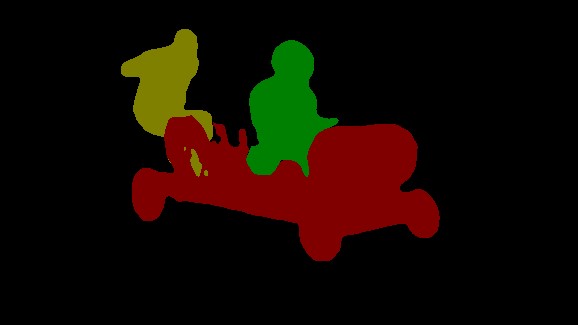}
	\end{minipage}
	}\hspace{-2mm}
	\subfloat{
	\begin{minipage}[c]{0.195\textwidth}
	\includegraphics[width=\textwidth]{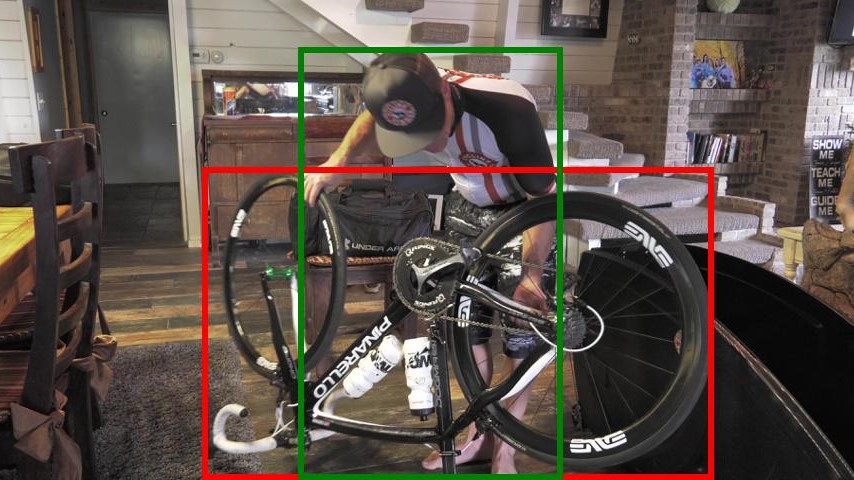}
    \includegraphics[width=\textwidth]{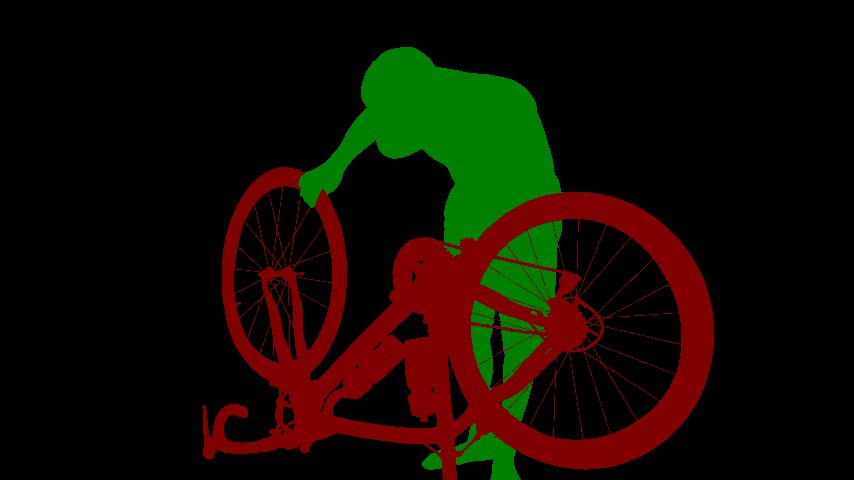}
    \includegraphics[width=\textwidth]{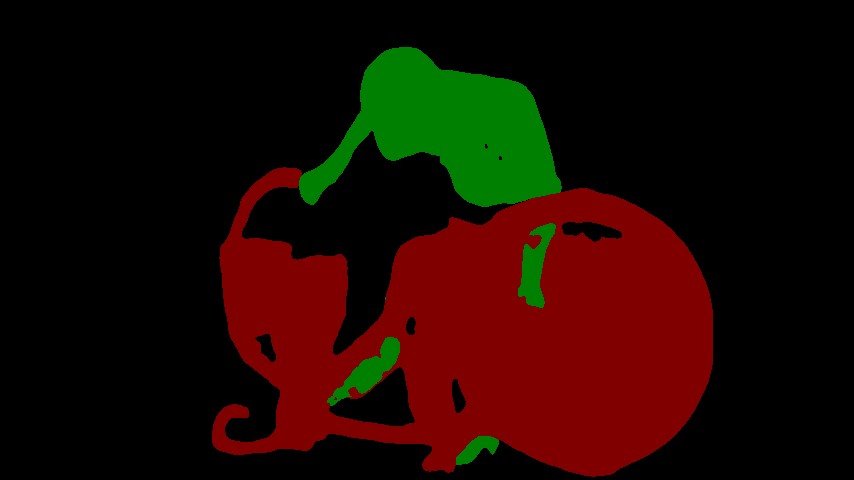}
	\end{minipage}
	}
    \caption{Additional qualitative results of our box to mask conversion network (third row) on DAVIS compared to the ground truth masks (second row).}
    \label{fig:qual_fig_sup}
\end{figure*}

\begin{figure*}
    \centering
    \subfloat{
    \begin{minipage}[c]{0.195\textwidth}
    \includegraphics[width=\textwidth]{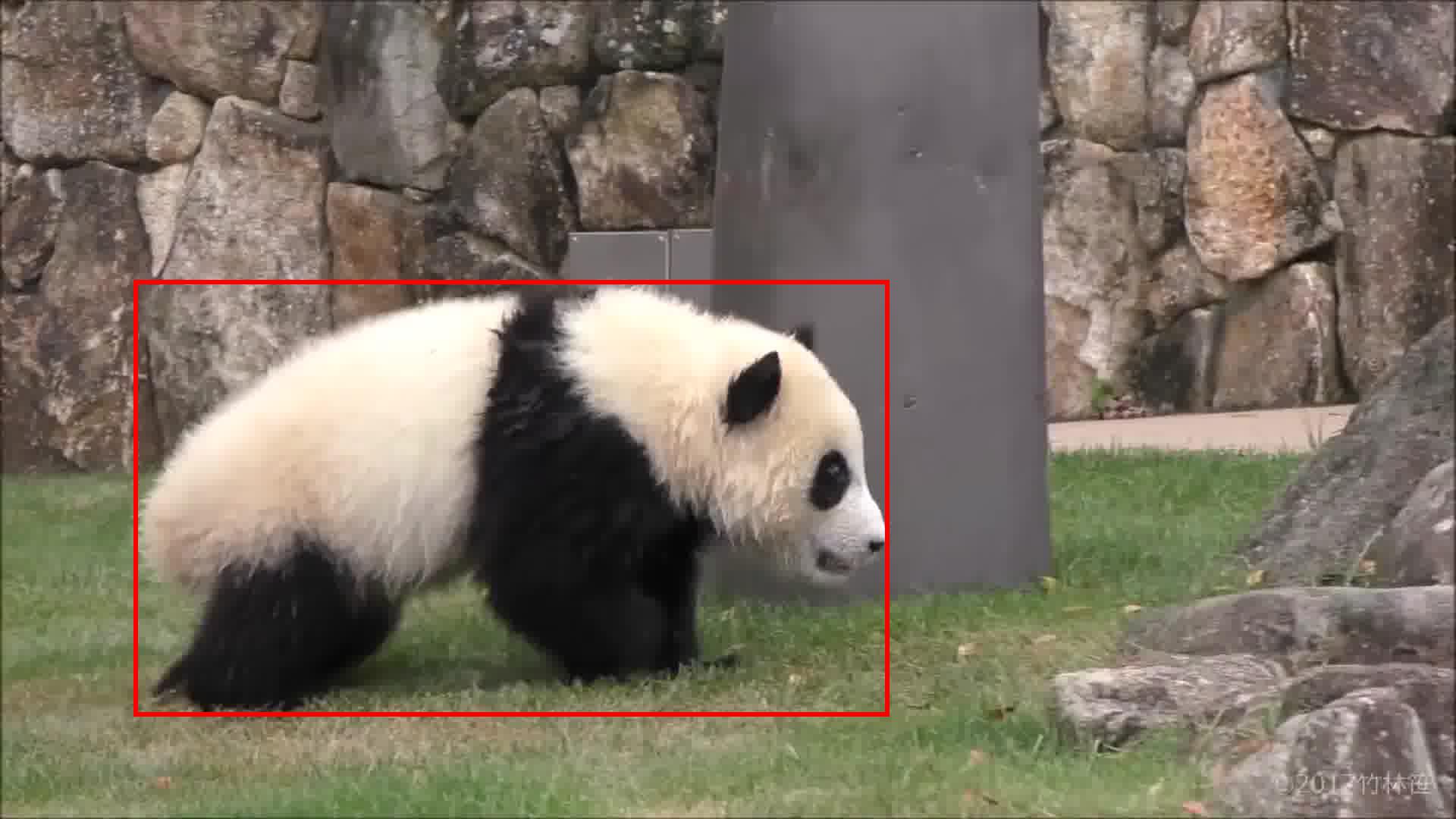}
    \includegraphics[width=\textwidth]{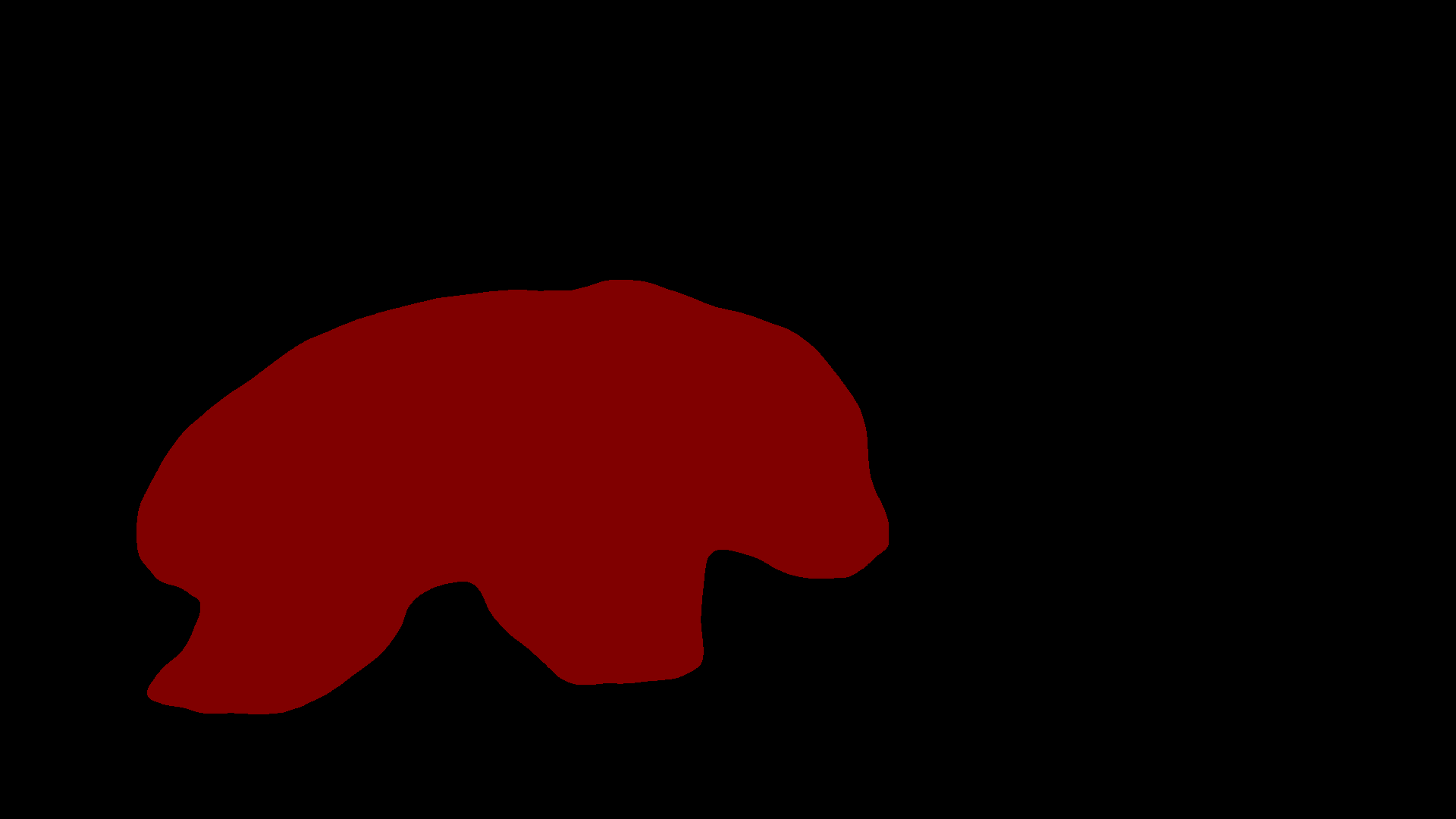}
	\end{minipage}
	}\hspace{-2mm}
	\subfloat{
    \begin{minipage}[c]{0.195\textwidth}
    \includegraphics[width=\textwidth]{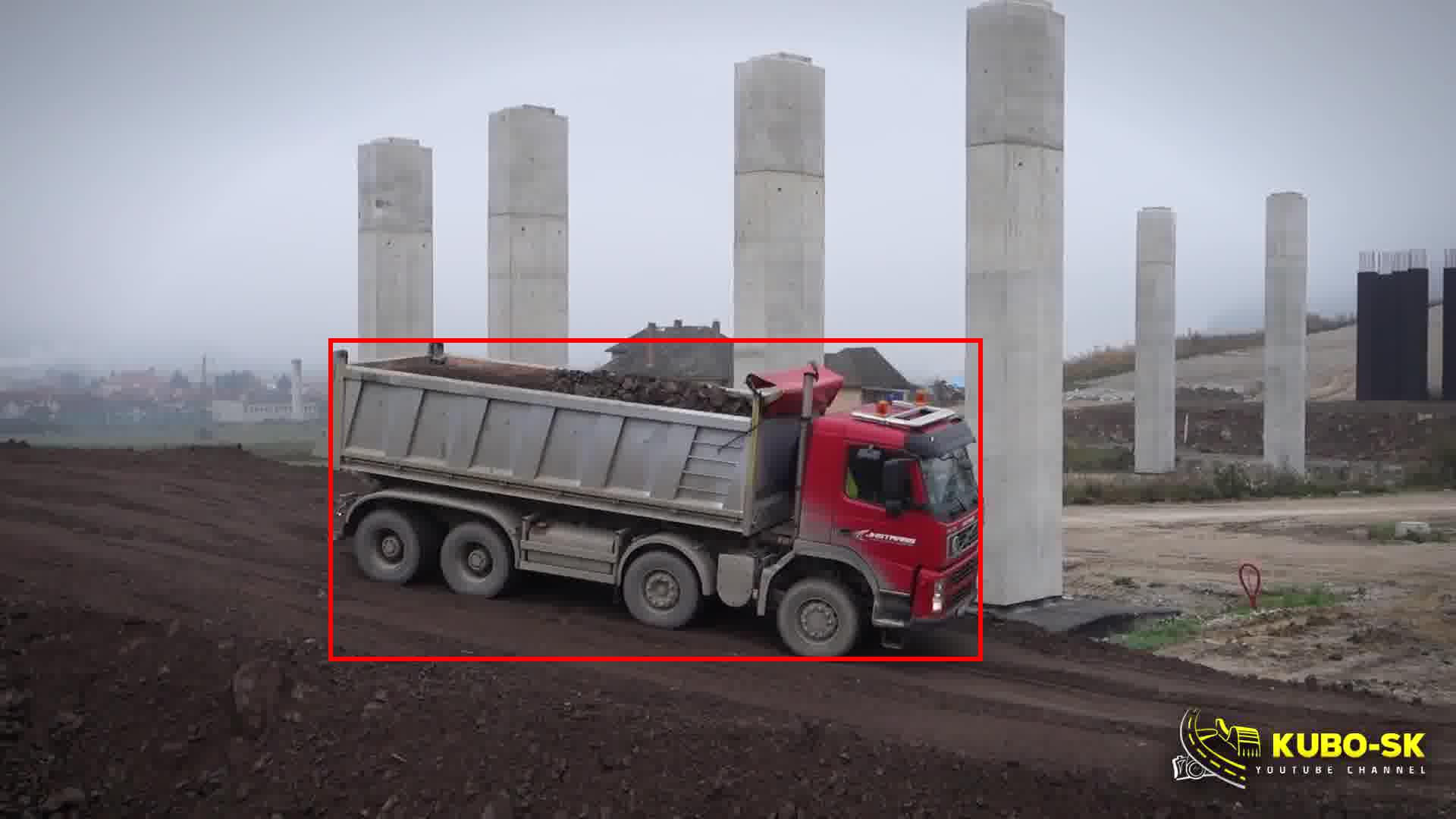}
    \includegraphics[width=\textwidth]{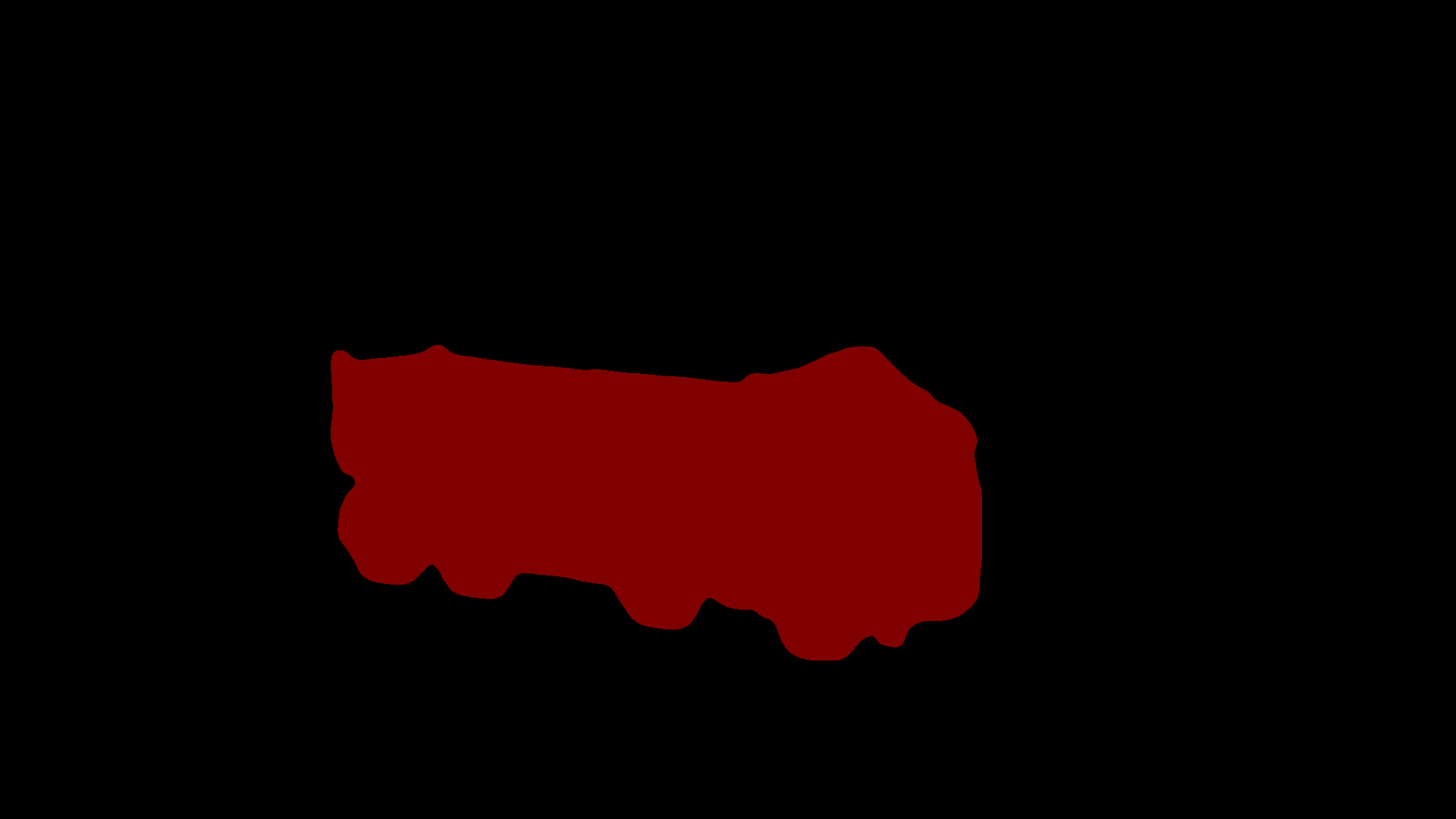}
	\end{minipage}
	}\hspace{-2mm}
	\subfloat{
    \begin{minipage}[c]{0.195\textwidth}
     \includegraphics[width=\textwidth]{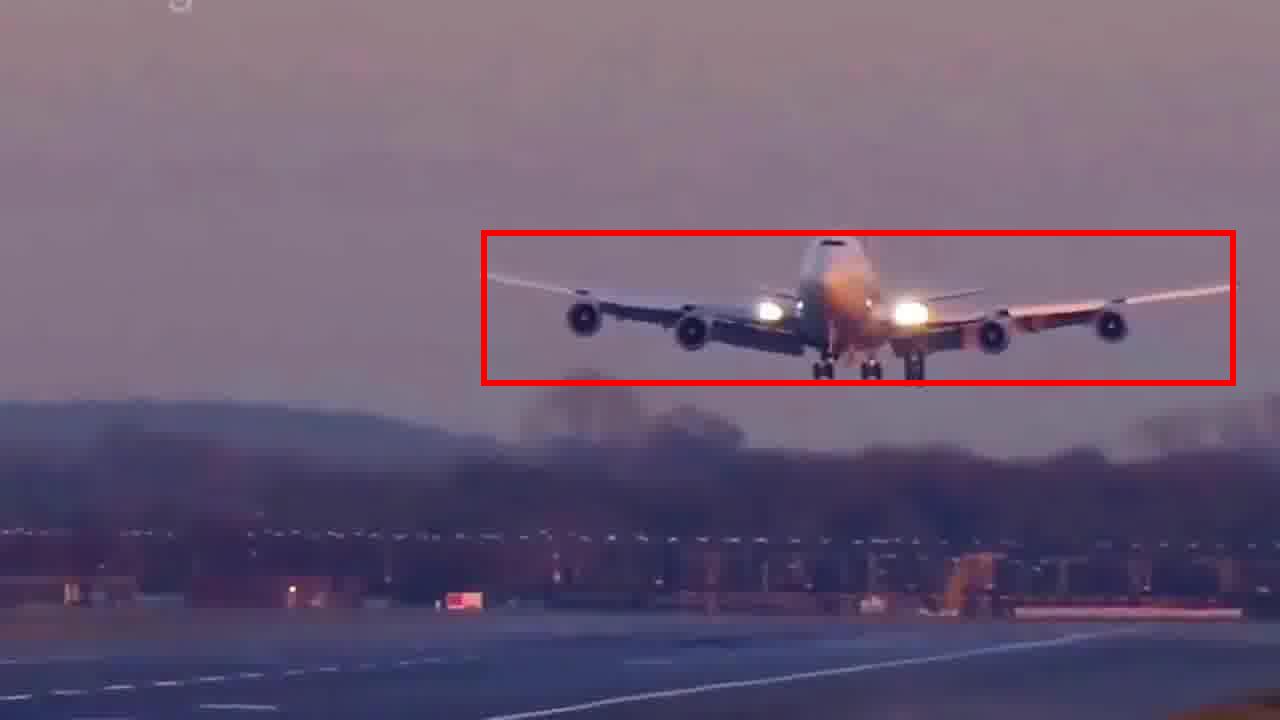}
     \includegraphics[width=\textwidth]{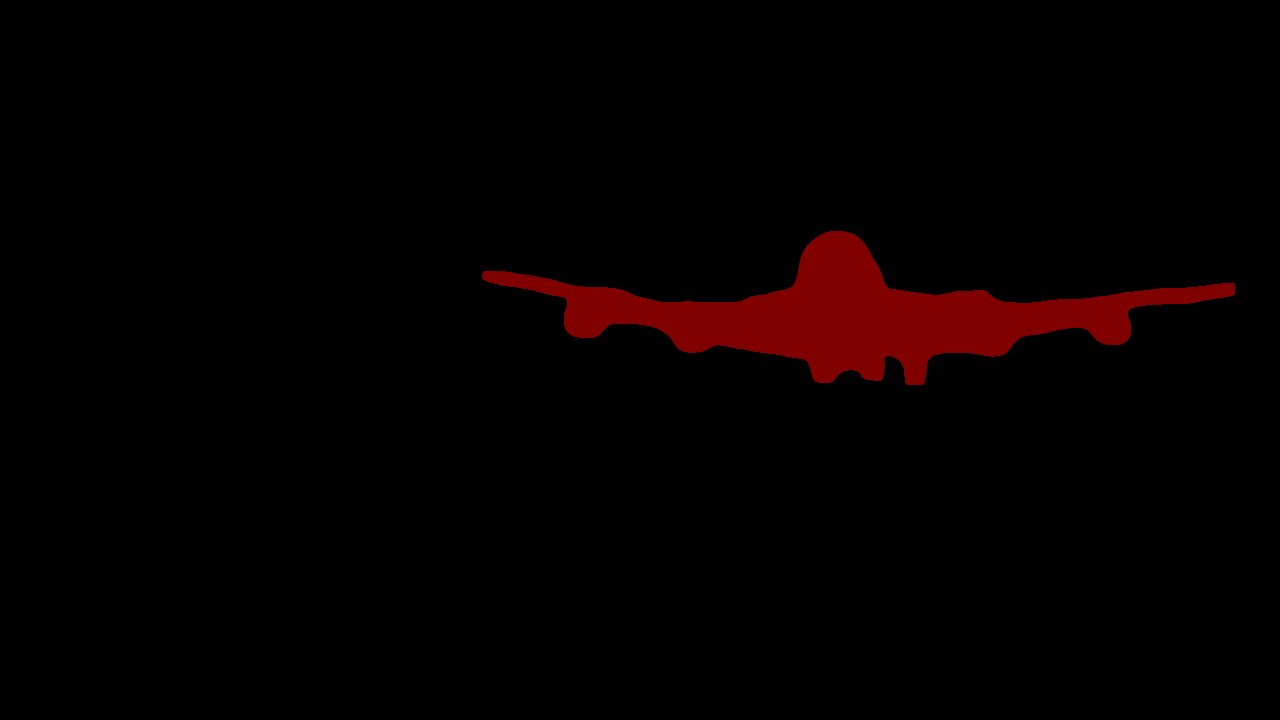}
	\end{minipage}
	}\hspace{-2mm}
	\subfloat{
	\begin{minipage}[c]{0.195\textwidth}
	\includegraphics[width=\textwidth]{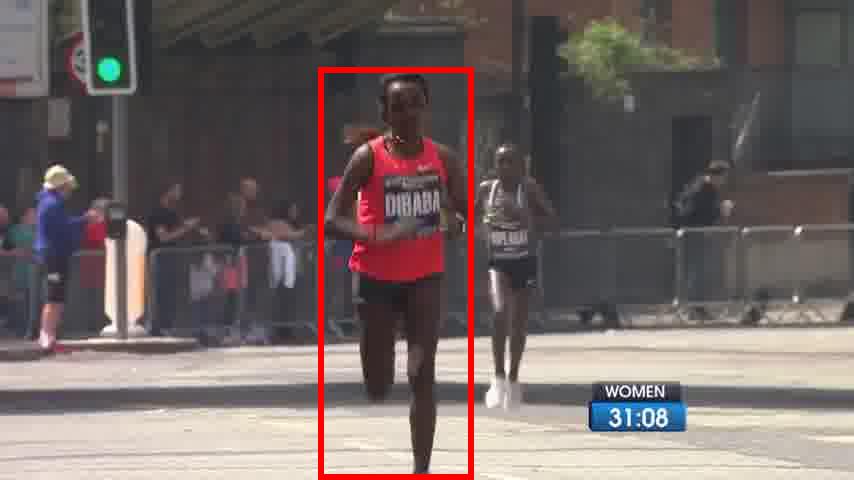}
    \includegraphics[width=\textwidth]{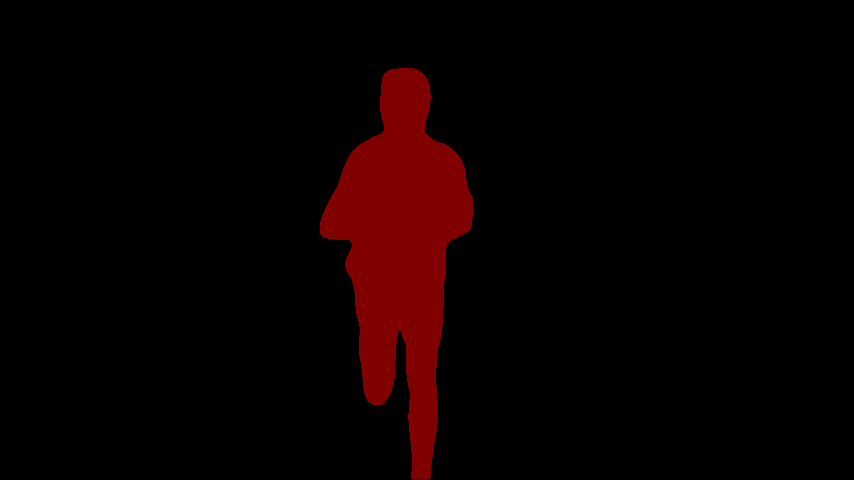}
	\end{minipage}
	}\hspace{-2mm}
	\subfloat{
	\begin{minipage}[c]{0.195\textwidth}
	\includegraphics[width=\textwidth]{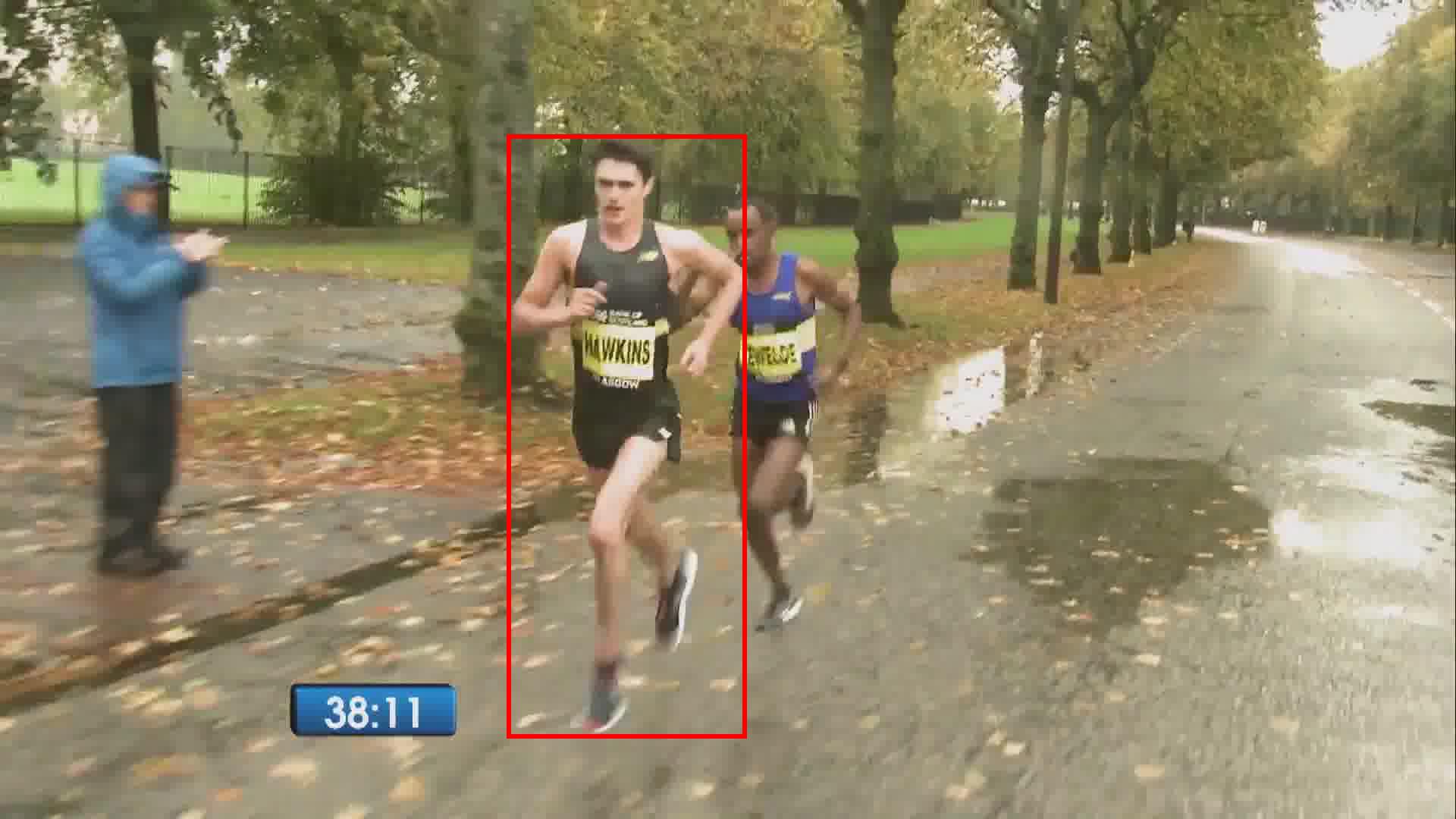}
    \includegraphics[width=\textwidth]{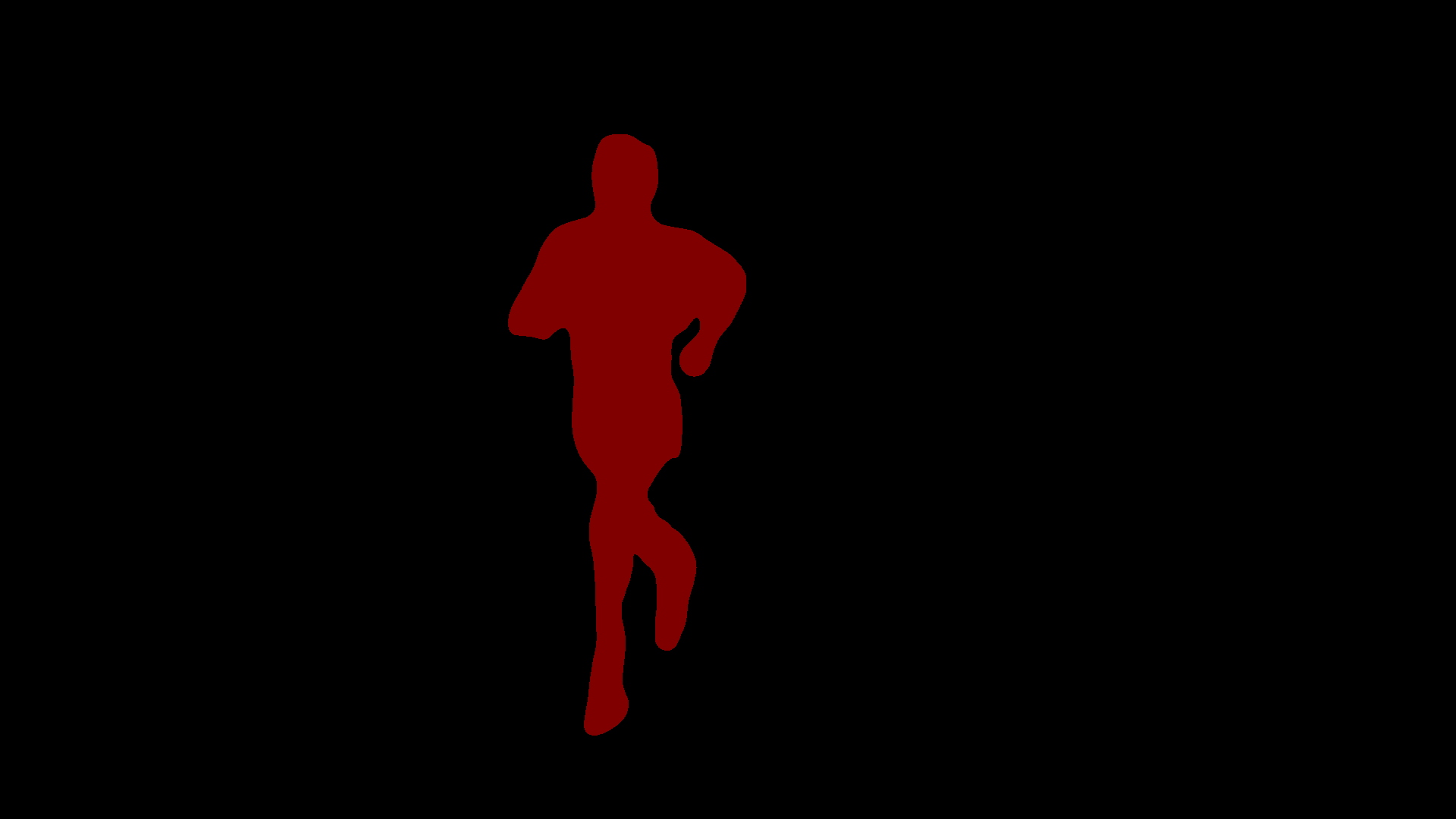}
	\end{minipage}
	}
    \caption{Additional qualitative results of our box to mask conversion network on GOT10k.}
    \label{fig:qual_fig_sup_got10k}
\end{figure*}

\section{Detailed analysis}
\label{sec:analysis}
Here, we provide a more detailed analysis of our approach for predicting object masks from bounding boxes in videos. 

\parsection{Impact of the number of steepest-descent iterations}
We perform 5 steepest-descent algorithm iterations during training in order to save training time and speed up convergence. However, there is no need to only iterate 5 times during inference. We perform experiments to analyse the impact of iterating more times and give results in Tab.~\ref{tab:Impact of the number of steepest-descent iterations}. Increasing the number of iterations from 5 to 15, the $\mathcal{J}$ score increases by $1.2$ on DAVIS2017 validation set. The performance of our approach saturates when iterating more than 15 times on YouTube-VOS and DAVIS.
\begin{table}[]
    \centering
    \begin{tabular}{l|cccc}
    \toprule
     Num. iterations    &  5 & 10 & 15 & 20\\
     \hline
     YT300     & 85.4 & 85.6 & 85.6 & 85.6 \\
     DAVIS2017 val & 80.0 & 80.9 & 81.2 & 81.2 \\
    \bottomrule
    \end{tabular}\vspace{1mm}
    \caption{Impact of the number of steepest-descent iterations. Results are shown in terms of Jaccard $\mathcal{J}$ index.}
    \label{tab:Impact of the number of steepest-descent iterations}
\end{table}

\parsection{Impact of the sample size} During training, we crop a patch 5 times larger than the ground truth bounding box to exploit background consistency. However, we find that it may be harmful to include too much unnecessary environment information. Therefore, we perform experiments to analyse the results if we crop a smaller patch. The size of the crop is set to different scaling factors relative to the object bounding box size. The results on YT300 and DAVIS2017 validation set are shown in Tab.~\ref{tab:Impact of the search area scale}. There is a significant improvement of $+1.2$ and $+0.7$ in $J$ score on YT300 and DAVIS when increasing the sample scale from 2 to 3, meaning including more background information leads to better results. We achieve best results when setting the sample scale as $3$ on YouTube-VOS and $4$ on DAVIS. A larger patch will contain too much noisy environment information, leading to degradation in performance. We set sample scale to 4 during inference in all our experiments.
\begin{table}[]
    \centering
    \begin{tabular}{l|cccc}
    \toprule
      Sample scale relative to object   &  2& 3 & 4 & 5\\
     \hline
     YT300     & 84.5 & 85.7 & 85.6 & 85.5 \\
     DAVIS2017 val & 80.2 & 80.9 & 81.2 & 80.8\\
    \bottomrule
    \end{tabular}\vspace{1mm}
    \caption{Impact of the image sample size relative to the object bounding box.  Results are reported in terms of Jaccard $\mathcal{J}$ score.}
    \label{tab:Impact of the search area scale}
\end{table}

\parsection{Impact of the inter-frame interval} Here, we analyse the inter-frame interval used for DAVIS and YouTube-VOS. Results in terms of Jaccard $\mathcal{J}$ score are shown in Tab.~\ref{tab:Impact of the inter-frame interval on DAVIS} and Tab.~\ref{tab:Impact of the inter-frame interval on YT300}. Compared to using consecutive frames, selecting every 5th frame gives an improvement of $+0.7$ in $\mathcal{J}$ score on DAVIS2017 validation set. The performance will decrease if we use larger inter-frame interval. While on YouTube-VOS, since the sequence is annotated every 5 frames, the minimal inter-frame interval that can be used is 5. There is no substantial difference of using different inter-frame intervals on YT300, so we generally set a  value 15 for other experiments.

\begin{table}[]
    \centering
    \begin{tabular}{l|cccc}
    \toprule
     Inter-frame interval    &  1 & 5 & 10 & 15\\
     \hline
     DAVIS2017 val & 80.5 & 81.2 & 80.9 & 80.5\\
    \bottomrule
    \end{tabular}\vspace{1mm}
    \caption{Impact of the inter-frame interval on DAVIS. Results are shown in terms of Jaccard $\mathcal{J}$ index.}
    \label{tab:Impact of the inter-frame interval on DAVIS}
\end{table}

\begin{table}[]
    \centering
    \begin{tabular}{l|ccccc}
    \toprule
     Inter-frame interval    & 5 & 10 & 15 & 20 & 25\\
     \hline
     YT300 & 85.6 & 85.7 & 85.6 & 85.6 &  85.5\\
    \bottomrule
    \end{tabular}\vspace{1mm}
    \caption{Impact of the inter-frame interval on YT300. Results are shown in terms of Jaccard $\mathcal{J}$ index.}
    \label{tab:Impact of the inter-frame interval on YT300}
\end{table}

\section{Additional inference details}
\label{sec:inf_details}
We describe details about how we annotate large-scale tracking datasets LaSOT and GOT10k. Generally, we use the same inference setting as used for YouTube-VOS, except for the number of frames.  We select 5 frames for each testing frame in order to save inference time and ensure the high performance at the same time. We annotate every frame of the sequence on GOT10K, while on LaSOT, we only annotate every 5th frame. LaSOT contains very long sequences and the objects generally move slowly. There is no need to annotate adjacent frames because they are highly correlated. Moreover, we only annotate up to 200 frames from a video sequence to avoid generating too much data for the same object.

\section{Detailed results}
\label{sec:detailed_results}

\parsection{GOT10K}
\begin{figure}[t]
	\centering%
	\includegraphics[trim = 0 0 0 0, width=\columnwidth]{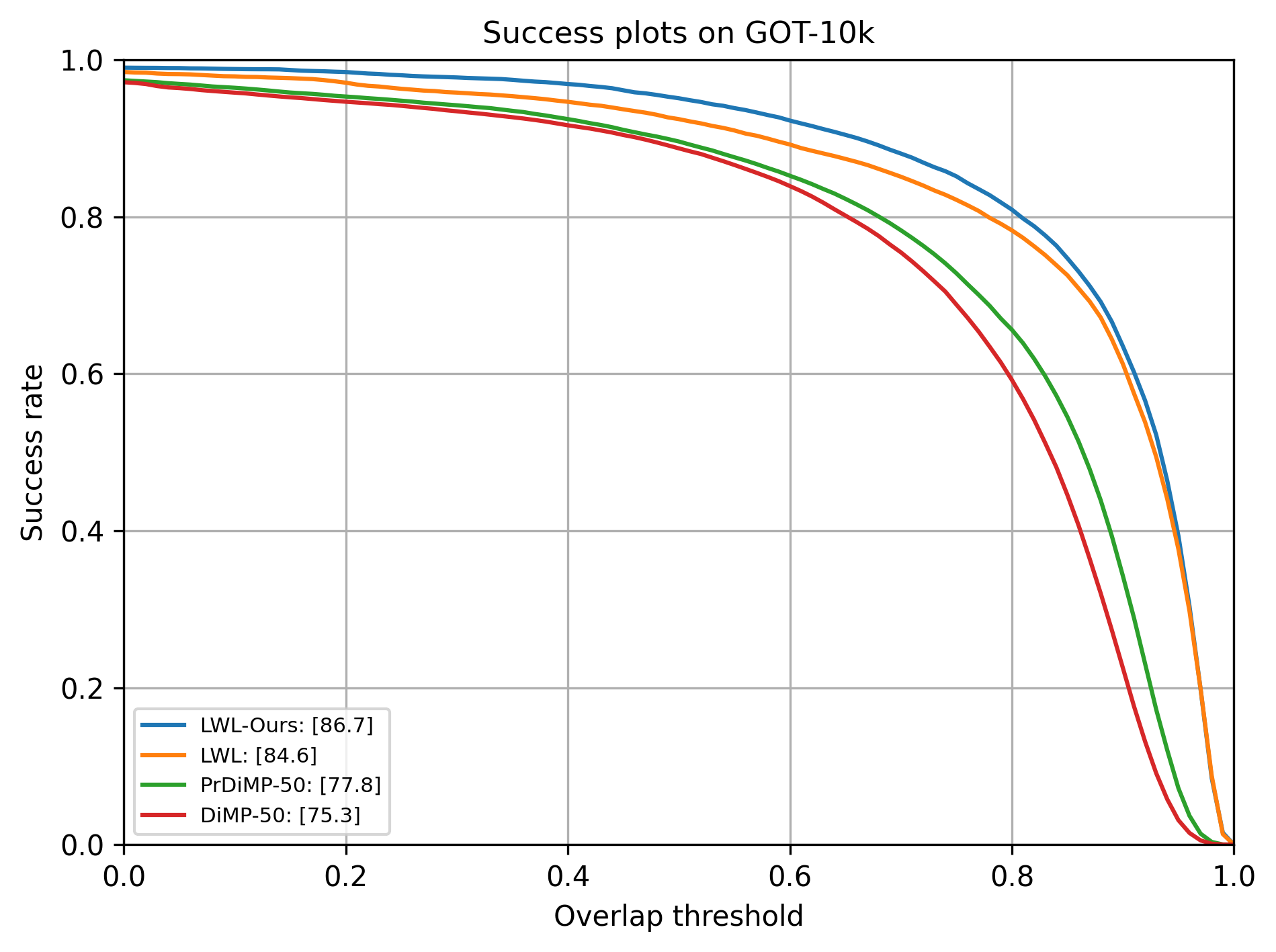}
	\caption{Success plots on the GOT10k validation set. The trackers are ranked using the average overlap (AO) score in percentage.}\vspace{0mm}
	\label{fig:got_suc}
\end{figure}
In this section, we provide success plots on the GOT10k validation set in Fig.~\ref{fig:got_suc}. The success plots are obtained using the Overlap Precision (OP) score. The OP score at a threshold $\tau$ denotes the fraction of frames in which the intersection-over-union (IoU) overlap between the tracker prediction and the ground truth box is greater than $\tau$. The OP scores for a range of thresholds in $[0, 1]$ are plotted to obtain the success plots. The trackers are ranked according the the average overlap (AO) score, which is computed as the average IoU overlap between the tracker prediction and the ground truth box over all frames in the dataset. The LWL model trained using our weakly annotated tracking data (LWL-Ours) obtains the best AO score, outperforming the standard LWL model $2.1\%$.

\parsection{YouTube-VOS} We compare our approach with state-of-the-art on YouTube-VOS 2018 validation set (see Tab.~\ref{tab:ytvos18}). LWL-Ours obtains comparable results with standard LWL and outperforms other VOS methods significantly. 
\begin{table}[h]
    \centering
    \scriptsize
    \setlength{\tabcolsep}{0.75mm}
    \begin{tabular}{lccccccc}
    \toprule
         &  \tabincell{c}{OSVOS \\ \cite{caelles2017one}} & \tabincell{c}{OnAVOS\\ \cite{voigtlaender2017online}} & \tabincell{c}{PreMVOS \\ \cite{luiten2018premvos}} & \tabincell{c}{SiamRCNN \\ \cite{voigtlaender2020siam}} & \tabincell{c}{STM \\\cite{oh2019video}}  & LWL & \textbf{LWL-Ours}\\
         \midrule
     $\mathcal{J}\&\mathcal{F}$ mean & 58.8 & 55.2 & 66.9 & 73.2 & 79.4 & 81.5  & 80.9\\
     \hline
     $\mathcal{J}_{seen}$ & 59.8 & 60.1 & 71.4 & 73.5 & 79.7 & 80.4 & 79.6\\
     $\mathcal{J}_{unseen}$ & 54.2 & 46.1 & 56.5 & 66.2 & 72.8 & 76.4 & 75.7\\
     \hline
     $\mathcal{F}_{seen}$ & 60.5 & 62.7 & - & - & 84.2 & 84.9 & 83.9\\
     $\mathcal{F}_{unseen}$ & 60.7 & 51.4 & - & - & 80.9 & 84.4 & 84.3\\
    \bottomrule
    \end{tabular}
    \vspace{1mm}
    \caption{Comparison on YouTube-VOS 2018 validation set.}
    \label{tab:ytvos18}\vspace{-5mm}
\end{table}

\section{Qualitative results}
\label{sec:qual}
We show more qualitative results on DAVIS and GOT10k in Fig.~\ref{fig:qual_fig_sup}. and Fig.~\ref{fig:qual_fig_sup_got10k}, respectively. Compared to the ground truth masks on DAVIS, our approach gives high-quality results for frames containing only a single object. For frames containing multiple objects, our approach can still delineate object boundary accurately if the overlapping problem is not severe.

\end{document}